\documentclass{article}

\PassOptionsToPackage{numbers, compress}{natbib}

\usepackage[dblblindworkshop, final]{neurips_2026}
\workshoptitle{Physical Understanding for Decision-Making}

\usepackage[utf8]{inputenc}
\usepackage[T1]{fontenc}
\usepackage{hyperref}
\usepackage{url}
\usepackage{booktabs}
\usepackage{amsfonts}
\usepackage{nicefrac}
\usepackage{microtype}
\usepackage{xcolor}
\usepackage{graphicx}
\usepackage{multirow}
\usepackage{float}
\usepackage{subcaption}
\usepackage{amsmath}
\usepackage{amssymb}

\title{Empirical Reset Pose Curriculum for Multi-Step Manipulation Task}

\author{%
  Wesley Wong \\
  Independent Researcher \\
  \texttt{wesleycwong@gmail.com} \\
}

\hypersetup{
    pdftitle={Empirical Reset Pose Curriculum for Multi-Step Manipulation Task},
    pdfauthor={Wesley Wong},
}

\begin{document}

\maketitle

\begin{abstract}
In reinforcement learning for robotics, particularly for long-horizon manipulation tasks, a fixed reset distribution forces the agent to keep re-traversing already mastered stages before ever reaching the point where it is currently struggling, an inefficient experience that compounds training's usual instability. We introduce a curriculum that instead adapts an environment's reset pose to the agent's own empirical sub-task performance, concentrating experience at its current bottleneck rather than repeatedly performing solved stages. It is \textbf{a seamless addition, requiring no change to the environment, reward, or learning algorithm}.
Using IsaacLab, we evaluate this curriculum across up to 16 seeds per environment on three multi-step manipulation tasks, reporting full results for Franka Cabinet, Franka Factory, and Franka Lift, and \textbf{11 ablation tests} on the curriculum.
Respectively, Cabinet, Factory, and Lift achieve Interquartile Mean (IQM) success rate absolute increases of \textbf{9.3} (70.7\% $\rightarrow$ 80.0\%), \textbf{2.2} (91.0\% $\rightarrow$ 93.2\%), and \textbf{62.7} (36.3\% $\rightarrow$ 99.0\%) points from their baselines. Factory's baseline is already tightly saturated near the ceiling, so this remaining gain is increasingly difficult to obtain.
The curriculum is also more reliable: on Lift, it converges on \textbf{14 of 16} seeds compared to the baseline's \textbf{7 of 16}, recovering runs that would otherwise collapse under the baseline.
\end{abstract}

\section{Keywords}
\label{sec:keywords}
\noindent\textbf{Sub-task}: intermediate milestone toward the overarching task. \\[1pt]
\textbf{Bottleneck}: earliest sub-task where progress is currently most stalled; the curriculum's replay target. \\[1pt]
\textbf{World State}: an environment's configuration (positions, velocities, orientations, and joint state) captured at sub-task completion and sufficient to reset the simulator back into it. \\[1pt]
\textbf{Progression}: live, per-environment tensor of sub-task completion and world state. \\[1pt]
\textbf{Success Buffer}: FIFO memory of world states saved at each sub-task's completion; sampled from. \\[1pt]
\textbf{Sampling Ratio}: fraction of resets that become \emph{curriculum episodes} rather than the environment's default (\emph{non-curriculum}) pose. \\[1pt]
\textbf{Difficulty}: complement of (EMA-smoothed) success rate. \\[1pt]
\textbf{Confident Sampling}: selecting the highest-difficulty sub-task via $\operatorname{argmax}$. \\[1pt]
\textbf{Exploratory Sampling}: sampling a sharpened difficulty distribution via $\operatorname{softmax}$. \\[1pt]
\textbf{ACES}: Adaptive Confidence-Exploration Sampler, continuously blending confident and exploratory sampling by margin. \\[1pt]
\textbf{Replay Specialization}: a narrow, non-generalizing solution learned from replayed resets that fails to transfer to the full task. \\[3pt]
Full definitions, and additional terms, are in Appendix~\ref{sec:s5-keywords}.

\section{Introduction}
Reinforcement learning (RL) is a leading approach for teaching physical robots complex, multi-step manipulation skills through direct interaction rather than hand-engineered control, but training remains slow, sample-inefficient, and unstable: identical setups trained with different random seeds can converge to markedly different final performance \cite{henderson2018deep}. This instability is often not a smooth spread of outcomes but a \emph{bimodal} split: most seeds converge to strong, comparable performance while a minority collapse to near-zero and never recover, a well-documented pattern in RL training rather than a sign of a broken task or environment. We return to this directly in Section~\ref{sec:results}, where it appears most clearly on Lift.
Curriculum learning is one of the most effective tools for addressing this instability, progressively introducing parts of a task rather than its full difficulty from the outset so an agent builds competence incrementally, converging faster, more reliably, and to a higher final performance than training on the complete task directly \cite{bengio2009curriculum}.
Existing curricula, however, largely reshape \emph{what} the agent is asked to do (adjusting rewards, goals, or environment parameters) rather than \emph{where} in a task it is currently struggling. An agent that has mastered a task's early stages but consistently fails at a later sub-task keeps re-experiencing the easy, already-solved stages before ever reaching its current bottleneck, spending training time exactly where it is least needed. This is not simply wasted time: repeatedly returning to already-solved stages also disrupts focus, pulling attention away from the transition where improvement is actually needed and compounding preexisting instability. \newline
We close this gap with the \textbf{Empirical Reset Pose Curriculum}: instead of always resetting to a fixed default pose, it lets an agent's own sub-task performance decide where training resumes, continually redirecting experience toward wherever it is currently struggling most.
To select this target, we design \textbf{ACES} (Adaptive Confidence-Exploration Sampler), which blends a decisive, confident choice with a more conservative, exploratory one based on how clearly a single bottleneck stands out from the rest \cite{de2018multi}.
Concretely, the curriculum tracks each sub-task's completion across training and resets a fraction of environments directly into states just prior to the identified bottleneck, drawn from a buffer of the agent's own previously successful attempts, requiring no change to the environment, reward, or learning algorithm. Resetting closer to a struggling transition also creates more chances to discover the rare action sequences that make it succeed, without re-earning the way through every earlier stage first. \newline
Because these policies will ultimately control a real robot's motors, training-time gains in success rate and consistency translate into fewer failed grasps, drops, or stalled manipulations after deployment, assuming successful sim-to-real transfer.
Across three multi-step manipulation tasks built on the Franka robot arm, the curriculum delivers a \textbf{9.3}-point IQM success-rate gain on Cabinet and a \textbf{2.2}-point gain on Factory despite an already-saturated, near-ceiling baseline that leaves little room to improve. On Lift, it delivers a \textbf{62.7}-point IQM gain and recovers \textbf{14 of 16} otherwise-collapsing seeds, compared to the baseline's \textbf{7 of 16} (Section~\ref{sec:results}). \newline
Section \ref{sec:related-work} situates our approach relative to prior work; Section \ref{sec:method} details the curriculum and ACES; Section \ref{sec:experimental-setup} discusses our setup. Sections \ref{sec:results}--\ref{sec:discussion} report and analyze results; Section \ref{sec:limitations} discusses limitations.

\section{Related work}
\label{sec:related-work}
The two key components of curriculum learning are (1) the part of training being adapted and (2) the signal that drives adaptation.
Task- and goal-adaptive curricula change \emph{what} the agent trains on: \citet{bengio2009curriculum} first formalized sampling from a distribution that changes over training, self-paced learning \cite{kumar2010self} let the learner decide which examples it is ready for from its own loss, and teacher-student curricula \cite{graves2017automated,matiisen2019teacher} adapt which task comes next from a learning-progress signal. Our curriculum instead keeps the task fixed and adapts only where an episode begins, using each sub-task's success rate directly.
Reset- and initialization-based methods are closest in spirit, manipulating the initial-state distribution directly: Reverse Curriculum Generation \cite{florensa2017reverse} expands a solvable region backward from a set of known goal states, though it strictly requires a determinisitic simulator or known backward dynamics to seed those initial states. Similarly, Go-Explore \cite{ecoffet2019go} manipulates state initialization by returning to previously visited states to keep exploring, relying heavily on deterministic simulator resets or complex goal-conditioned policies to navigate back. Reset-free RL \cite{zhou2024reset} instead asks whether resets are needed at all. None reallocate among a fixed, empirically populated set of reset poses using per-sub-task success, as ours does.
Environment-adaptive methods change the environment itself: domain randomization \cite{tobin2017domain}, PAIRED \cite{dennis2020emergent}, and POET \cite{wang2019paired} broaden environment variability at the frontier of agent capability, but require generating new environments, which is considerably more complex than adapting a fixed environment's reset distribution.
To the best of our knowledge, ours is the first curriculum that (1) adapts the initial-state distribution of an otherwise fixed task, (2) using an agent's own recorded, per-sub-task success statistics, and (3) without external demonstrations, known goal geometry, or a generative model of goals or environments.

\section{Method}
\label{sec:method}
\subsection{Overview}
\label{sec:overview}
We now describe the curriculum's components: how it optimizes an environment's reset poses during training to improve agent learning.
The overall task is broken into sub-tasks marking milestones toward the end goal.
At every step, the curriculum logs each sub-task's completion across all environments and saves the world state at the moment of completion; this \emph{progression} is updated continuously as a live representation of training, while a separate \emph{success buffer} retains a memory of past successful world states for future resets.
The progression yields a per-sub-task difficulty score, computed only from \emph{non-curriculum episodes} (episodes reset to the environment's default pose) to avoid \emph{eval contamination} biasing the signal the curriculum itself depends on.
This difficulty score is converted into a sampling distribution by \textbf{ACES} (\textbf{A}daptive \textbf{C}onfidence-\textbf{E}xploration \textbf{S}ampler), the sampling algorithm we introduce in this work (Section \ref{sec:distribution-algorithms}). As environments terminate or truncate, a fraction become \emph{curriculum episodes}, sampling a target sub-task from this distribution and drawing a world state from that sub-task's buffer to reset into.

\begin{figure}[H]
\centering
\includegraphics[width=0.8\textwidth]{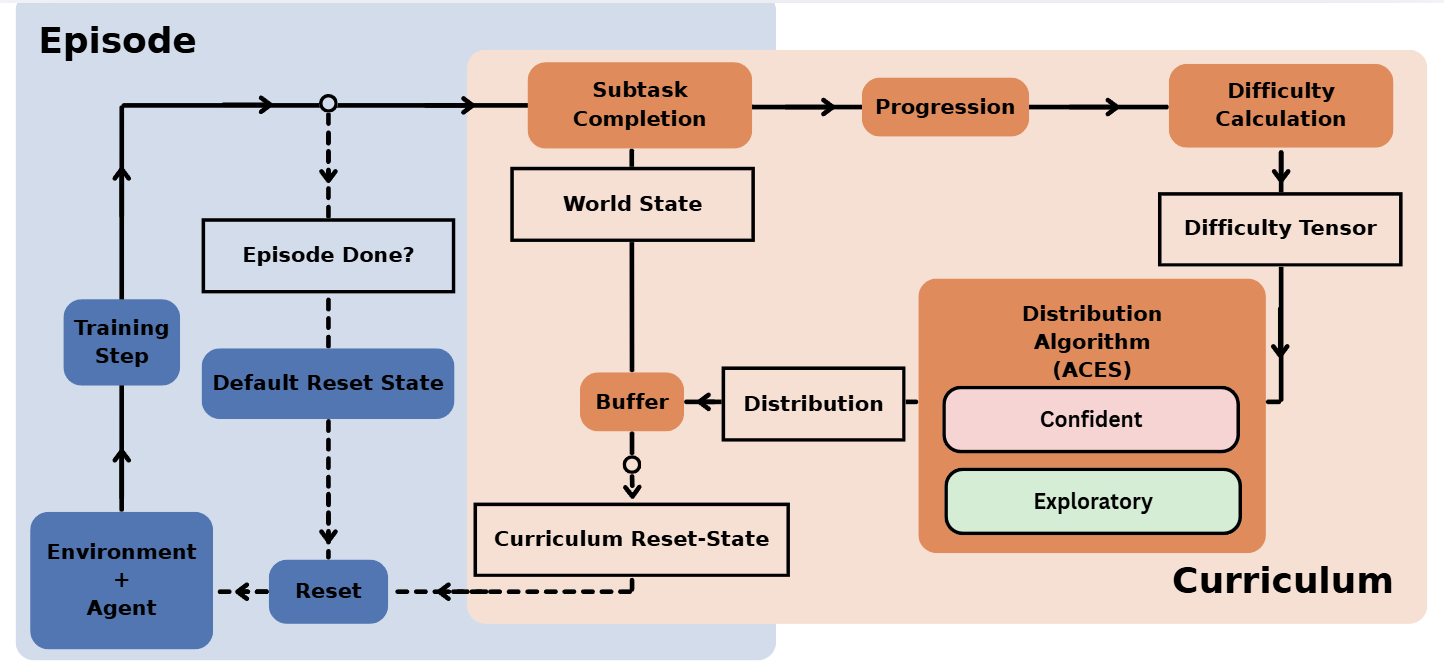}
\caption{A concise flowchart of the curriculum within a single training episode. Blue (\emph{Episode}) is the standard RL train-step/reset loop; orange (\emph{Curriculum}) is our contribution. Dashed arrows: the reset path; dots: branch points.}
\label{fig:overview-diagram}
\end{figure}

\subsection{Sub-task and world representation}
\label{sec:world-state-modeling}
A sub-task is an intermediate milestone toward the main goal, most naturally a discrete action, though it can also be a continuous quantity such as a target distance. Each sub-task's completion is a boolean stored in a tensor of shape $[N,S]$ ($N$ environments, $S$ sub-tasks), computed every step by a sub-task function (e.g.\ reaching within a threshold, or grasping an object).
A world state captures the environment configuration at the moment a sub-task completes (positions, velocities, orientations, and joint state, relative to the environment origin), sufficient for the simulator to reset back into it exactly.
It is stored in two places: the \emph{progression}, a live tensor $P \in \mathbb{R}^{N \times S \times (W+1)}$ used to calculate sub-task difficulty, where $P_{n,i,0} \in \{0,1\}$ is environment $n$'s completion of sub-task $i$; and the \emph{success buffer}, a FIFO tensor of shape $[S,B,W]$ ($B$ the buffer-size hyperparameter, $W$ the world representation size) that pushes a world state the moment $P_{n,i,0}$ flips from 0 to 1, the buffer the curriculum ultimately samples world states from.
Exact per-field storage conventions (e.g.\ which quantities are omitted for fixed objects) are in Appendix~\ref{sec:s3-hyperparams}.
\subsection{Difficulty estimation}
\label{sec:difficulty-estimation}
Let $\mathcal{E} \subseteq \{1, \dots, N\}$ denote environments currently running a non-curriculum episode (Section \ref{sec:overview}).
Each sub-task's difficulty is estimated from the \emph{progression's} batch success rate of these environments (mean completion $P_{n,i,0}$ across $\mathcal{E}$, denoted $u_i$), EMA-smoothed into a persistent estimate $r$ with coefficient $\alpha$ (the \textit{success-rate EMA coefficient}, Table \ref{tab:curriculum_hyperparameters}) to filter batch-to-batch noise \cite{sutton1998reinforcement}, then complemented into a failure rate $d \in [0,1]^S$:
\begin{equation}
u_i = \frac{1}{|\mathcal{E}|} \sum_{n \in \mathcal{E}} P_{n,i,0}, \qquad r \leftarrow (1 - \alpha)\, r + \alpha\, u, \qquad d = 1 - r
\end{equation}
To reveal the earliest bottleneck, we compute the gap between each sub-task's difficulty and the one before it, shifted non-negative with a small $\epsilon$ so uniform gaps do not collapse to zero:
\begin{equation}
g_i = d_i - d_{i-1}, \quad d_0 := 0, \qquad \bar g = g - \min(g) + \epsilon
\end{equation}
$\bar g \in \mathbb{R}_{\geq 0}^S$ feeds the distribution algorithm described next.
\subsection{Distribution algorithms}
\label{sec:distribution-algorithms}
We developed and evaluated a variety of sampling algorithms, resulting in \textbf{ACES} (\textbf{A}daptive \textbf{C}onfidence-\textbf{E}xploration \textbf{S}ampler), which blends a sharpened exploratory sampler ($\operatorname{softmax}$) with a confident sampler ($\operatorname{argmax}$), weighted by how confidently the gap tensor $\bar g$ identifies a single bottleneck \cite{de2018multi}.
The exploratory component sharpens $\bar g$ by exponent $p$ (the \textit{difficulty exponent}) before $\operatorname{softmax}$, treating the bottleneck with less confidence:
\begin{equation}
\sigma = \operatorname{softmax}\!\left(\bar g^{\,p}\right)
\end{equation}
Separately, we normalize $\bar g$ into a confidence tensor $c$, and take its one-hot argmax as the confident component $h$ (e.g.\ $h=[0,1,0,0]$ if sub-task 2 is most confident):
\begin{equation}
c = \frac{\bar g}{\sum_i \bar g_i}, \qquad h_i = \begin{cases} 1 & i = \operatorname{argmax}_i\, c_i \\ 0 & \text{otherwise} \end{cases}
\end{equation}
The \emph{margin} $m$ (the gap between $c$'s two largest values, $c_{(1)}$ and $c_{(2)}$) captures how confidently a single sub-task stands out: a large margin means one sub-task is clearly the bottleneck (sparse $c$), a small margin means several are similarly difficult (dense $c$).
This determines the \emph{blend} weight $\beta$ via the \textit{confidence margin} hyperparameter $M_g$:
\begin{equation}
m = c_{(1)} - c_{(2)}, \qquad \beta = \operatorname{clamp}\!\left(\frac{m}{M_g},\ 0,\ 1\right)
\end{equation}
As the margin grows past $M_g$, $\beta \to 1$ and ACES becomes purely confident selection; as it shrinks toward 0, $\beta \to 0$ and ACES becomes purely exploratory.
The final curriculum distribution is this blend, renormalized to sum to one:
\begin{equation}
\mathcal{D} = \frac{(1-\beta)\,\sigma + \beta\,h}{\sum_i \left[(1-\beta)\,\sigma + \beta\,h\right]_i}
\end{equation}
For efficiency, $\mathcal{D}$ is recomputed with fixed probability $\rho=0.10$ (the \textit{distribution update probability}) per step while the curriculum is enabled, and deterministically every $1/\rho=10$ steps while disabled.
\subsection{Curriculum replay}
The curriculum \textit{replays} by sampling from $\mathcal{D}$ and resetting environments to a saved world state.
When an environment terminates (task completion or a safety violation) or truncates (timeout), a sampling ratio $\gamma \in [0,1]$ (the \textit{curriculum sampling ratio}) governs what fraction of resets on that step become curriculum episodes rather than resetting to the default pose.
Each curriculum episode draws a target sub-task $i \sim \mathcal{D}$, then a world state uniformly from that sub-task's buffer slice, and resets into it; the progression is cleared for any environment that has just finished an episode.
\section{Experimental setup}
\label{sec:experimental-setup}
\subsection{Environments}
We now describe the environments, metrics, and ablations used to evaluate the curriculum specified above.
We evaluate three environments built on a 7-DOF Franka arm using IsaacLab's \cite{mittal2025isaaclab} example configurations, training with PPO \cite{schulman2017proximal} (an actor-critic method \cite{barto1983neuronlike}) via RSL-RL \cite{schwarke2025rslrl} (Cabinet, Lift) or RL-Games (Factory); full PPO/network hyperparameters are in Appendix~\ref{sec:s3-hyperparams}. \newline
\textbf{Franka Cabinet} (2500 iter, 16384 envs, Direct): reach within 20\,cm $\to$ reach within 10\,cm of the drawer handle $\to$ open the drawer 20\,cm $\to$ open it 38\,cm [\textit{final goal: open 39\,cm}]. \newline
\textbf{Franka Factory} (200 iter, 512 envs, Direct): align the nut with the bolt $\to$ engage the threads (gap $<0.5\times$ thread pitch) $\to$ rotate the nut down 0.75\,mm from engagement [\textit{final goal: the same threshold as this last sub-task, see Section \ref{sec:design-guidelines}}]. \newline
\textbf{Franka Lift} (3000 iter, 4096 envs, Manager-Based): reach within 10\,cm of the cube $\to$ grasp it (within 2\,cm, gripper clamped) $\to$ lift it 10\,cm $\to$ hold it within 5\,cm of the elevated target pose. \newline

\begin{table}[H]
\centering
\caption{Reset curriculum hyperparameters, shared across all three environments.}
\label{tab:curriculum_hyperparameters}
\small
\begin{tabular}{lc}
\toprule
\textbf{Hyperparameter} & \textbf{Value} \\
\midrule
Success buffer size & 64 \\
Difficulty exponent ($p$) & 2 \\
Curriculum sampling ratio ($\gamma$) & 0.3 \\
Reset pose domain randomization & 0.02$^{*}$ \\
Success-rate EMA coefficient ($\alpha$) & 0.05 \\
Confidence margin ($M_g$) & 0.10 \\
Distribution update probability ($\rho$) & 0.10 \\
\bottomrule
\end{tabular} \\[2pt]
{\footnotesize $^{*}$Factory uses 0.00 (no reset-pose domain randomization), since its sub-task geometry is pose-sensitive (Section~\ref{sec:limitations})}
\end{table}

\subsection{Baselines and metrics}
\label{sec:baselines}
Success is not computed uniformly across environments: Cabinet and Lift log a continuous completion ratio (e.g.\ Cabinet's drawer position normalized to its 0.39\,m full-open threshold, clamped to $[0,1]$), while Factory logs a strict binary pass/fail across all three conditions. All three are logged to TensorBoard and, like every success-rate metric here, computed only from non-curriculum episodes to avoid eval contamination.
Cabinet and Lift use \textbf{16 seeds (42--57)}; \textit{Factory uses 8 seeds (42--49), reduced from 16 given its substantially longer training time.}
All seeds are deterministic, verified against an unmodified, curriculum-free build and against $\gamma=0$ (Appendix~\ref{sec:s4-discussion}).
Beyond overall success rate, we track per-sub-task success (curriculum vs.\ non-curriculum), the resulting success gap (Figure \ref{fig:replay-vs-eval}), and the blend/margin diagnostics of Section \ref{sec:distribution-algorithms}, with the full diagnostic and RL metric set in Appendix~\ref{sec:s2-diagnostics}.
\subsection{Ablation studies}
We trained eleven ablations on Cabinet, each a single deviation from the default configuration, together probing how the curriculum should commit to a bottleneck, how much training to redirect, when to start, and what state to replay (Table \ref{tab:curriculum_hyperparameters}, same 16384-environment, 2500-iteration budget).
We discuss three key tests in depth below: \textbf{Confident} (Argmax-only), \textbf{Exploratory} (Softmax-only), and \textbf{Delay} (curriculum disabled for the first 15\% of training). The remaining eight are reported in full in Appendix~\ref{sec:s1-ablations}; none reached significance against baseline except Sampling Ratio 0.5, significantly \textit{worse} ($p=0.044$, Table \ref{tab:math-analysis}).
\section{Results}
\label{sec:results}
\subsection{Overall performance}
\begin{figure}[H]
\centering
\begin{subfigure}[t]{0.28\textwidth}
\includegraphics[width=\textwidth]{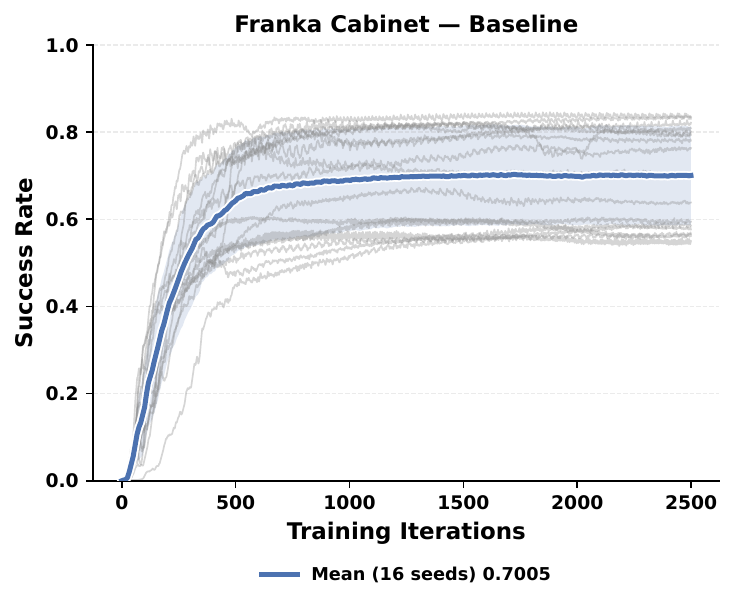}
\end{subfigure}\hfill
\begin{subfigure}[t]{0.28\textwidth}
\includegraphics[width=\textwidth]{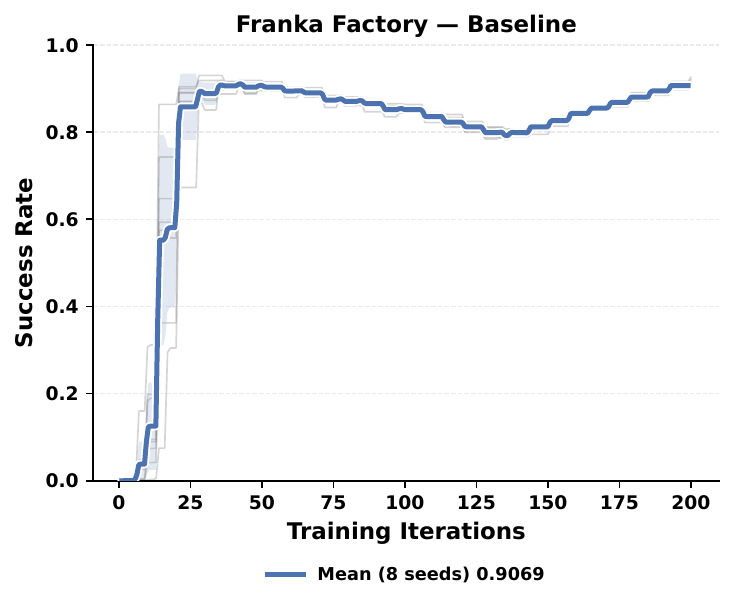}
\end{subfigure}\hfill
\begin{subfigure}[t]{0.28\textwidth}
\includegraphics[width=\textwidth]{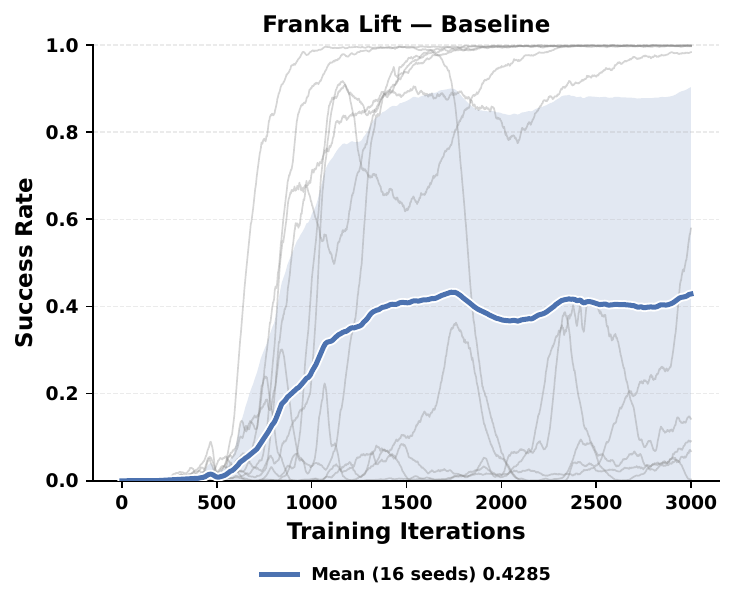}
\end{subfigure}

\begin{subfigure}[t]{0.28\textwidth}
\includegraphics[width=\textwidth]{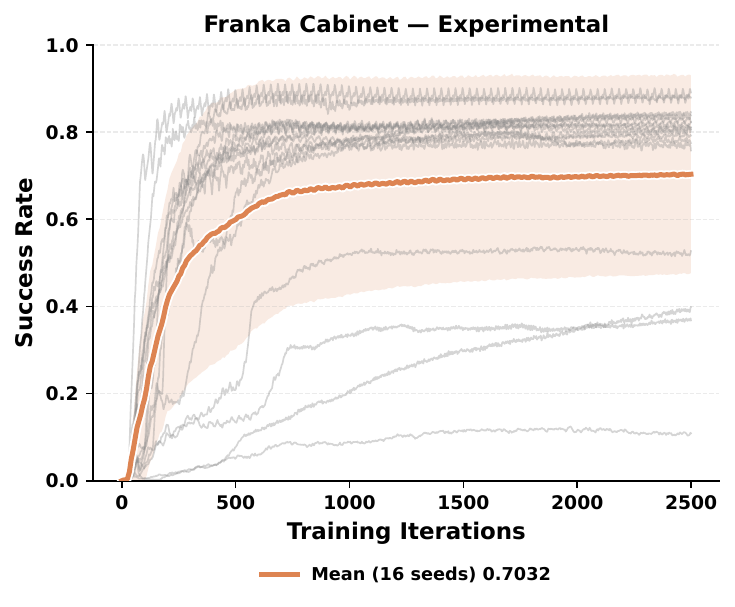}
\end{subfigure}\hfill
\begin{subfigure}[t]{0.28\textwidth}
\includegraphics[width=\textwidth]{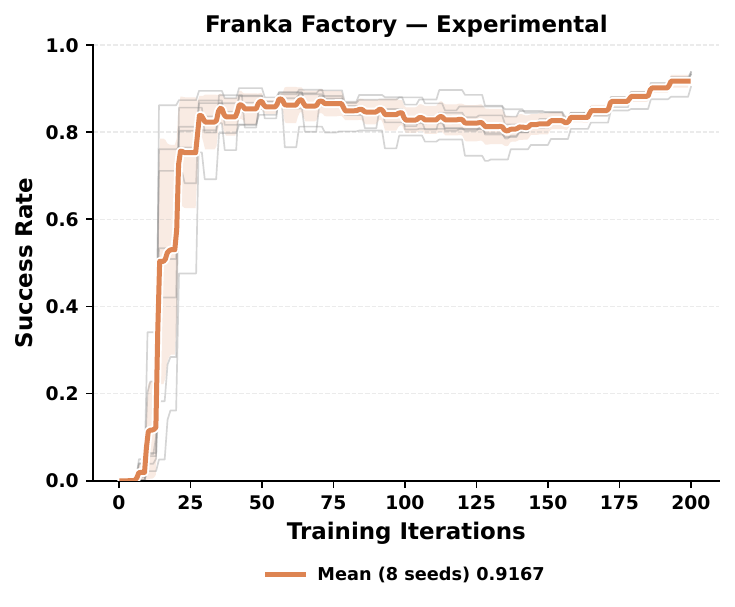}
\end{subfigure}\hfill
\begin{subfigure}[t]{0.28\textwidth}
\includegraphics[width=\textwidth]{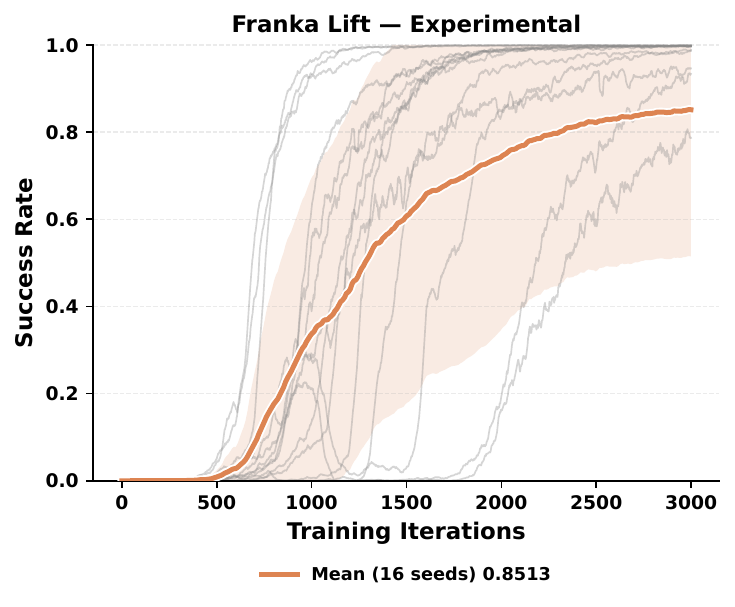}
\end{subfigure}
\caption{Mean success rate over training, baseline (top) vs.\ curriculum (bottom), left to right: Cabinet, Factory, Lift. Gray: individual seeds; shaded: $\pm1$ std; bold: mean. Lift is smoothed with a 21-step trailing moving average; Factory tracks a 9-step rolling maximum per seed to filter logging noise. Reported final values are unsmoothed; full detail in Appendix~\ref{sec:s3-hyperparams}.}
\label{fig:three-environments}
\end{figure}

Figure \ref{fig:three-environments} gives the complete overview: baseline vs.\ curriculum learning curves for all three environments. Table \ref{tab:math-analysis} reports full statistics for every comparison below, merging descriptive and paired-test results.

\begin{table}[H]
\centering
\caption{Overall performance (Cabinet, Factory, Lift) and Cabinet ablations (vs.\ baseline, below the rule). IQM~\cite{agarwal2021deep}/Mean/Median/Std.\ describe that row's own condition (Min/Max in Appendix~\ref{sec:s1-ablations}); Cliff's $\delta=2P(X{>}Y)-1$, the 95\% bootstrap CI on $P(X{>}Y)$ ($B=10{,}000$ paired resamples~\cite{agarwal2021deep}), and Wilcoxon $p$ are paired comparisons against baseline, omitted for baseline rows. Confident/Exploratory/Delay are discussed below (Section \ref{sec:ablation-studies-heading}); the rest are in Appendix~\ref{sec:s1-ablations}.}
\label{tab:math-analysis}
\footnotesize
\setlength{\tabcolsep}{4.5pt}
\begin{tabular}{llccccccc}
\toprule
\textbf{Env.} & \textbf{Condition} & \textbf{IQM} & \textbf{Mean} & \textbf{Median} & \textbf{Std.} & \textbf{95\% CI} & \textbf{Cliff's $\delta$} & \textbf{Wilcoxon $p$} \\
\midrule
\multirow{2}{*}{Cabinet} & Baseline & 0.7072 & 0.7004 & 0.7321 & \textbf{0.1137} & --- & --- & --- \\
 & Experimental & \textbf{0.7995} & \textbf{0.7032} & \textbf{0.8073} & 0.2269 & $[0.375, 0.875]$ & $+0.250$ & 0.5619 \\
\midrule
\multirow{2}{*}{Factory} & Baseline & 0.9101 & 0.9102 & 0.9110 & \textbf{0.0097} & --- & --- & --- \\
 & Experimental & \textbf{0.9323} & \textbf{0.9308} & \textbf{0.9327} & 0.0119 & $[0.625, 1.000]$ & $+0.750$ & \textbf{0.0156} \\
\midrule
\multirow{2}{*}{Lift} & Baseline & 0.3629 & 0.4310 & 0.1222 & \textbf{0.4741} & --- & --- & --- \\
 & Experimental & \textbf{0.9895} & \textbf{0.8535} & \textbf{0.9957} & \textbf{0.3366} & $[0.438, 0.875]$ & $+0.375$ & \textbf{0.0110} \\
\midrule
\multicolumn{2}{l}{Confident} & 0.6892 & 0.6484 & 0.7369 & 0.2027 & $[0.188, 0.688]$ & $-0.125$ & 0.5282 \\
\multicolumn{2}{l}{Exploratory} & 0.6272 & 0.5721 & 0.6459 & 0.2601 & $[0.250, 0.750]$ & $+0.000$ & 0.3225 \\
\multicolumn{2}{l}{Delay} & 0.7647 & \textbf{0.7198} & 0.7692 & \textbf{0.1613} & $[0.313, 0.813]$ & $+0.125$ & 0.4037 \\
\bottomrule
\end{tabular}
\end{table}
\textit{Cabinet, Lift, and the ablations use $n=16$ seeds (42--57); Factory uses $n=8$ (42--49), given its longer training time. Factory's Mean/Median/Std.\ are each seed's peak success rate over its final 25 logged steps (Figure \ref{fig:three-environments}'s caption, Appendix~\ref{sec:s3-hyperparams}).}

\begin{figure}[H]
\centering
\begin{subfigure}[b]{0.28\textwidth}
\includegraphics[width=\textwidth]{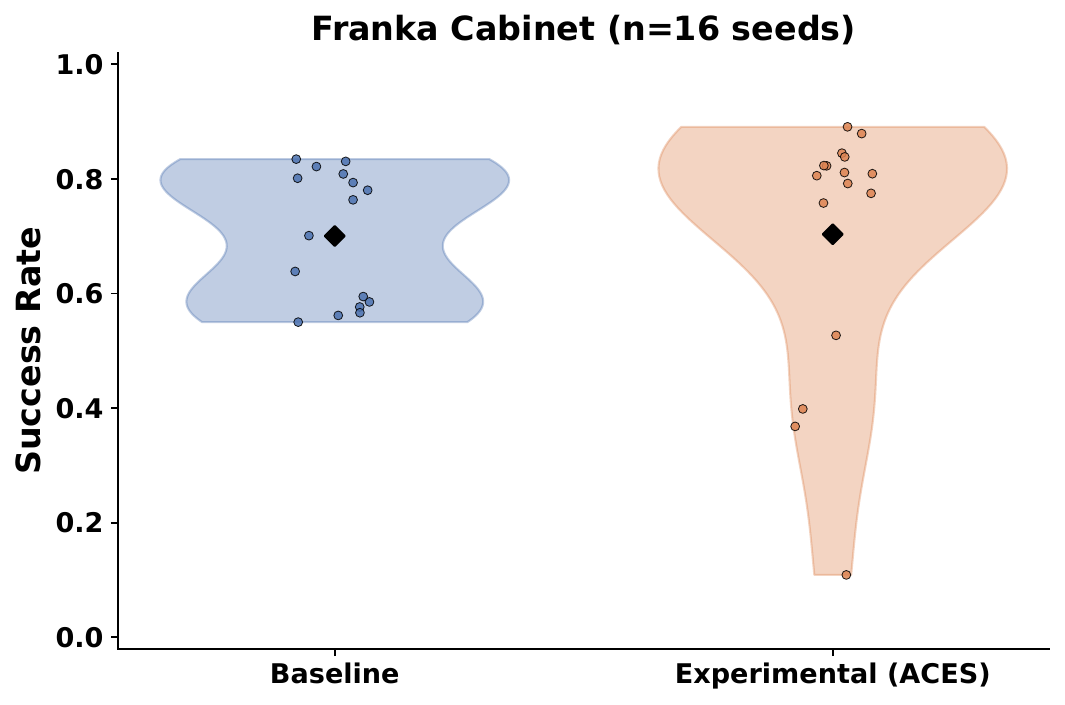}
\caption{Cabinet}
\label{fig:cabinet-violin}
\end{subfigure}\hfill
\begin{subfigure}[b]{0.28\textwidth}
\includegraphics[width=\textwidth]{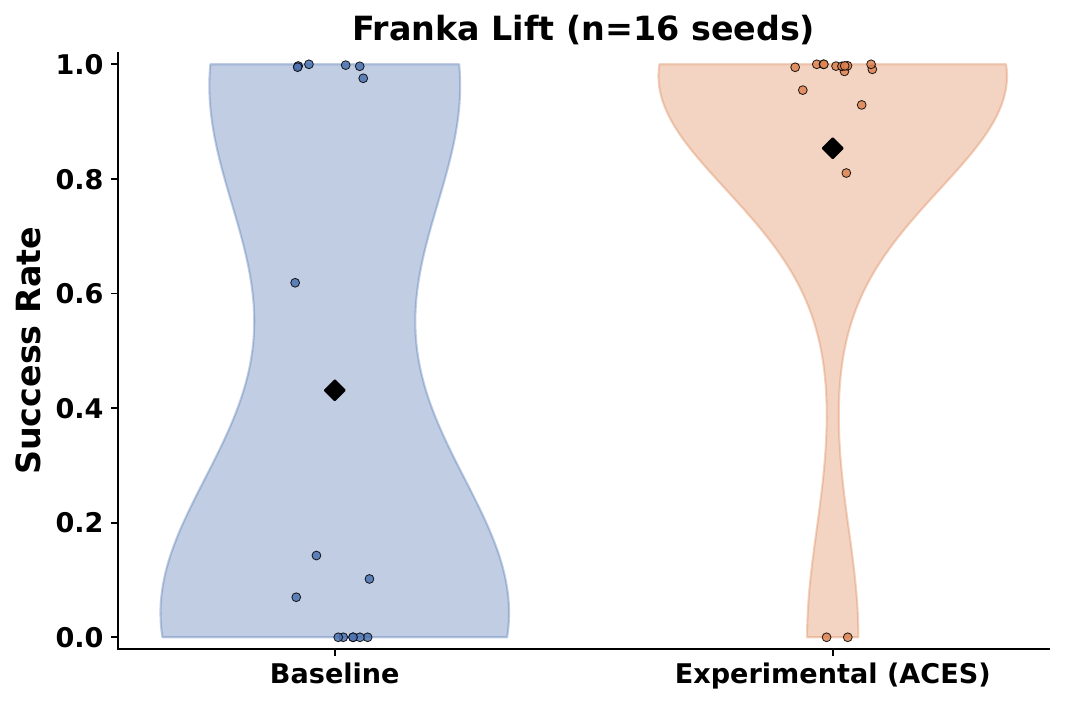}
\caption{Lift}
\label{fig:lift-violin}
\end{subfigure}\hfill
\begin{subfigure}[b]{0.28\textwidth}
\includegraphics[width=\textwidth]{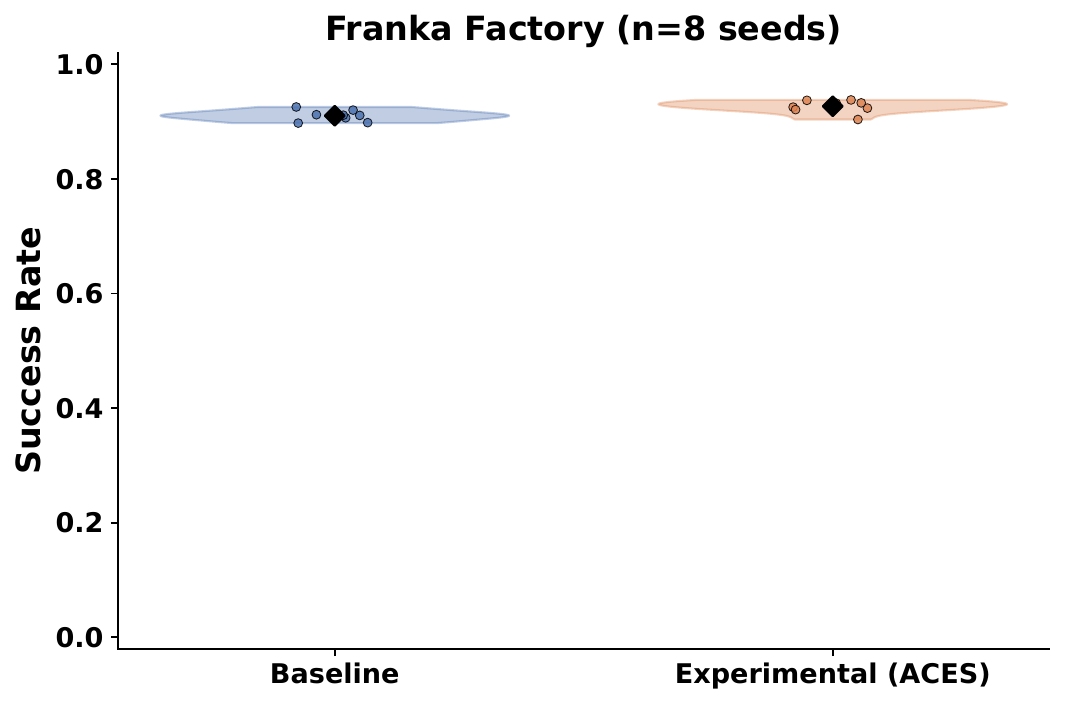}
\caption{Factory}
\label{fig:factory-violin}
\end{subfigure}
\caption{Final success-rate distribution, baseline vs.\ curriculum, violin plots with individual seeds (jittered) and group means (black diamonds).}
\label{fig:violin-panel}
\end{figure}

Figure \ref{fig:cabinet-violin} makes Cabinet's split outcome visible: ACES's high-performing cluster sits higher and tighter than baseline's, but a tail of catastrophic outliers (min $10.9\%$ vs.\ baseline's $55.0\%$) pulls the raw mean nearly flat even as median and IQM both improve. This is the outlier sensitivity that keeps \textbf{Cabinet from reaching classical significance} despite IQM, median, and the bootstrap CI on $P(X{>}Y)$ (Table \ref{tab:math-analysis}, entirely above 0.5) agreeing on a positive shift (Section \ref{sec:instability-replay-specialization}).
\textbf{Lift, by contrast, does reach significance} ($p=0.011$): Figure \ref{fig:lift-violin}'s strongly bimodal shape (Section \ref{sec:lift-extended-training-discussion}) reflects a curriculum that recovers three of five seeds that were still near-zero partway through training to $0.93$--$1.00$ by the end, where the baseline recovers none of nine comparably-stuck seeds past $15\%$\label{sec:lift-extended-training}. The final tally reflects the same gap: the curriculum converges on 14 of 16 seeds (final success above $15\%$) against 7 of 16 for the baseline.
\textbf{Factory shows gains despite an already-saturated baseline.} Figure \ref{fig:factory-violin}'s tight clustering reflects a strict, binary success condition (Section \ref{sec:baselines}) that leaves baseline little room to improve ($0.9102\pm0.0097$), yet the curriculum still raises the mean to $0.9308\pm0.0119$, \textbf{reaching significance} at the smallest sample size in this study ($n=8$, $p=0.016$), with \textbf{7 of 8 seeds improving}. Performance this close to ceiling is difficult to move.

\subsection{Confident, Exploratory, and Delay}
\label{sec:ablation-studies-heading}
Of the eleven ablations trained (Section \ref{sec:experimental-setup}), three are most insightful for how the curriculum should allocate experience: Confident and Exploratory test how aggressively it should commit to one bottleneck versus spread across several, and justify blending rather than picking one strategy; Delay tests whether early bottleneck estimates are reliable enough to target immediately. The remaining eight are in Appendix~\ref{sec:s1-ablations}.

\begin{figure}[H]
\centering
\begin{subfigure}[t]{0.48\textwidth}
\includegraphics[width=\textwidth]{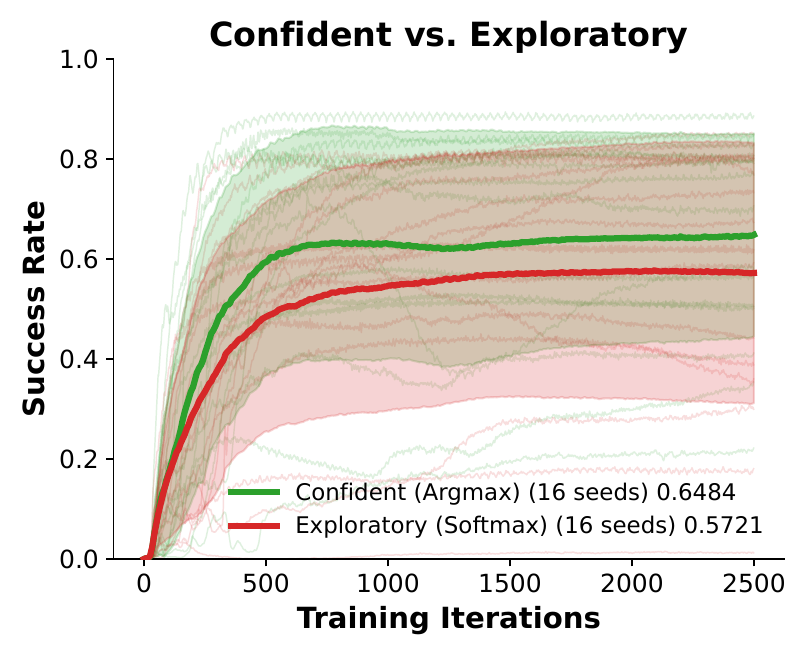}
\caption{Confident vs.\ Exploratory}
\label{fig:confident-vs-exploratory}
\end{subfigure}\hfill
\begin{subfigure}[t]{0.48\textwidth}
\includegraphics[width=\textwidth]{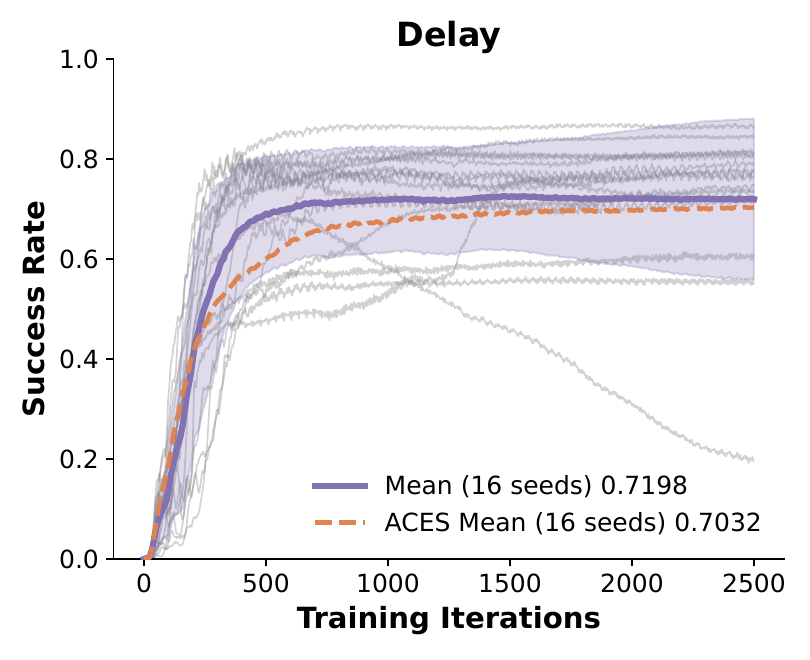}
\caption{Delay}
\label{fig:aces-vs-delay}
\end{subfigure}
\caption{Franka Cabinet, $n=16$. (a) Confident ($\operatorname{argmax}$-only) vs.\ Exploratory ($\operatorname{softmax}$-only, $p=2$). (b) Delay, with ACES's mean overlaid (dashed) for comparison (Table \ref{tab:math-analysis}). Shaded bands: $\pm1$ std, clipped to $[0,1]$.}
\label{fig:cabinet-panel}
\end{figure}
Pure confident and exploratory sampling both underperform the full ACES blend on mean ($0.648$ and $0.572$ vs.\ $0.703$), \textbf{direct evidence that blending the two, rather than committing to either alone, is what the curriculum benefits from}.
\label{sec:delay-ablation}\textbf{Delay is best read as a more stable variant of ACES, not a strictly better one}: its mean ($0.720$) is higher and its std ($0.161$) is the lowest of any curriculum-enabled condition, but it loses to ACES on IQM ($0.765$ vs.\ $0.800$) and wins only $31.25\%$ of paired seed comparisons (Section \ref{sec:delayed-onset}).

\section{Discussion}
\label{sec:discussion}
\subsection{Algorithm design: landing on ACES}
We now interpret these results, beginning with ACES's design.
An ideal sampler would always target the current empirical bottleneck exclusively, but difficulty estimation is never perfect, especially early in training; so the two core samplers apply the success-rate signal differently: the exploratory sampler discounts it, while the confident sampler applies it directly.
The exploratory sampler's softmax is deliberately less confident, spreading sampling probability across sub-tasks in proportion to their difficulty rather than committing to one; this makes it well suited to a dense difficulty distribution, where several sub-tasks are similarly hard and no single one clearly dominates, since committing to just one under that ambiguity would be premature.
The confident sampler produced higher-performing runs than the exploratory sampler, yet also more catastrophic ones, locking onto a sub-task prematurely whenever a marginally larger difficulty appeared as reliable as a clearly dominant one; it nonetheless performed best whenever it locked onto the correct bottleneck, indicating the failure mode was premature commitment under ambiguous signal, not a flaw in the strategy itself. \newline
ACES's continuous blend improves on two previous algorithms evaluated first, a fixed exploratory-to-confident scheduler and a discrete margin threshold (ACTS), both too blunt about when to switch; detailed in Appendix~\ref{sec:s4-algo-history}.
This blend genuinely shifts during training on Cabinet (Figure \ref{fig:blend-weight-cabinet}, Appendix~\ref{sec:s2-diagnostics}): exploratory dips are most densely concentrated in the first few hundred iterations, when per-sub-task difficulty estimates are noisiest and an exploratory sampler is most appropriate, before the blend settles into a mostly-confident regime (time-averaged $\beta \approx 0.90$) for the remainder of training as those estimates stabilize.
\label{sec:why-sharpening}A related finding: sharpening the exploratory sampler ($p=2$ vs.\ 1) improved both stability and performance (full analysis in Appendix~\ref{sec:s4-stability-analysis}).

\subsection{Instability \& replay specialization}
\label{sec:instability-replay-specialization}
Despite these design choices, a few runs still fail to converge (\emph{catastrophic runs}, whose success rate remains near zero), visible as the low-lying seed lines in Figure~\ref{fig:three-environments}'s Cabinet panel and the catastrophic tail in Figure~\ref{fig:cabinet-violin}. Both trace back to the same tension: concentrating experience on the bottleneck helps resolve it, but too much comes at the cost of the full-task exposure that keeps behavior general.

\textbf{Curriculum hyperparameter interactions.} Parallel environment count, buffer size, and sampling ratio jointly determine the freshness and diversity of sampled reset poses: too small or stale a buffer floods the curriculum with niche poses, and too high a ratio overpowers general-task learning. 0.3 works well; 0.5 degrades performance sharply (Table \ref{tab:math-analysis}, Appendix~\ref{sec:s1-ablations}).

\textbf{Replay specialization.} We hypothesize, though do not fully confirm, that these failures reflect \emph{replay specialization}: susceptibility to niche sub-task solutions that fail to transfer to the full task, amplified (not introduced) by the curriculum letting the agent learn within isolated sub-tasks, producing catastrophic failures in already-vulnerable environments (Cabinet and Lift alike). Appendix~\ref{sec:s-pure-subtask-replay} shows targeted resetting, not replay alone, drives the benefit.

\textbf{Diagnostic evidence.} This theory is supported, not proven, by three diagnostic points confirmed in Appendix~\ref{sec:s-good-bad-seed}: well-performing seeds converge to a lower, more stable \emph{reset pose deviation} while poor-performing seeds had noisier, elevated deviations; \emph{entropy} is lower in catastrophic seeds without translating to success; and \emph{value loss} spikes then declines for well-performing seeds but rises gradually, sometimes late, for catastrophic ones.
Per-sub-task success rates were similar across both groups: replays themselves succeeded either way, but that learning did not transfer back to the default, non-curriculum starting pose.

\subsection{Delayed onset and Lift recovery: two paths to reliability}
\label{sec:delayed-onset}
The Delay ablation (Section \ref{sec:delay-ablation}) produced an unplanned finding: withholding curriculum replay for training's first 15\% yields the lowest-variance curriculum-enabled configuration tested, raising the mean by \textbf{1.9} (70.04\% $\rightarrow$ 71.98\%), though it appears to lower max performance.
We hypothesize this works because two sources of early instability (noisy difficulty estimates and premature exposure to later sub-task resets, Section \ref{sec:instability-replay-specialization}) are both worst at the very start of training, exactly what the window Delay removes. This was not a primary hypothesis and was only tested on Cabinet, but a simple, essentially free change is one of the more promising directions this work opens up; full mechanistic reasoning is in Appendix~\ref{sec:s4-delayed-onset-full}. \newline
\label{sec:lift-extended-training-discussion}A central question for interpreting these results: does the curriculum merely reach a fixed ceiling faster, or unlock a better outcome entirely? Over the full 3000-iteration Lift run, double IsaacLab's default, the baseline's catastrophic runs remained low-performing and even its successful runs converged lower than the curriculum's, supporting the latter. (The default training length is contained within the first half of the 3000 iteration training)
The seven recovered seeds (Section \ref{sec:lift-extended-training}) suggest the curriculum can eventually unlock convergence that would otherwise fail outright, a distinct function beyond converging faster.

\subsection{Design guidelines: choosing sub-tasks and world representations}
\label{sec:design-guidelines}
Sub-tasks are an empirical interface for locating progress and failure, not a claim they capture the task's true decomposition. Effective ones correspond to a recognizable skill, are well spread out and achievable, and, recommended though not required, place the final sub-task one step short of full completion (e.g.\ Cabinet's 38\,cm sub-task vs.\ 39\,cm success) rather than identical to it; Factory does not follow this (Appendix~\ref{sec:s3-hyperparams}), full reasoning in Appendix~\ref{sec:s4-design-guidelines-subtasks}. Tasks that cannot be split into two or more sub-tasks cannot use our curriculum.
For world representation, positions and joints should always be preserved, but including joint velocities \emph{lowered} Cabinet performance (Velocity ablation, Appendix~\ref{sec:s1-ablations}), likely an uncaused signal from replaying a velocity the agent did not itself cause (full reasoning in Appendix~\ref{sec:s4-design-guidelines-world-rep}). Beyond these choices, the curriculum changes no reward or termination logic, only the sub-task functions and world-state fields above: a small implementation surface for the gains reported in Section~\ref{sec:results}.
\section{Limitations}
\label{sec:limitations}
\textbf{Scope.} Results span 16 seeds per configuration for Cabinet and Lift, 8 for Factory (Section \ref{sec:baselines}), across three environments on a common Franka arm, tested only with IsaacLab. Evaluation runs throughout training rather than as a separate held-out benchmark. Sensor-based environments, other tasks or robot types, and sim-to-real deployments were not tested. Only a no curriculum baseline was used, but benchmarking against alternative curriculums is the next immediate step.

\textbf{Replay specialization.} The curriculum produces success rates of up to $89.1\%$ on its best Cabinet seeds, but roughly a fifth become catastrophic, converging to about half the typical success rate (Section \ref{sec:instability-replay-specialization}).

\textbf{Implementation.} Using the curriculum requires manually decomposing the task into sub-tasks, access to reset the simulator into arbitrary saved states, and may require re-tuning hyperparameters for a different environment count or sampling ratio; the extra bottleneck-detection and sampling computation should modestly increase training time if unoptimized. Reset-pose domain randomization is disabled for Factory (Table \ref{tab:curriculum_hyperparameters}): jittering a reset mid-thread-engagement can offset the nut from the bolt's threads by a fraction of the thread pitch, an invalid rather than coherent engagement. A subtle IsaacLab reimplementation pitfall is in Appendix~\ref{sec:s-determinism}.

\section{Conclusion}
This paper introduced a curriculum that resets an agent to its own current bottleneck using sub-task progress rather than a hand-designed schedule.
On Cabinet, the curriculum raises IQM success rate by 9.3 points, though catastrophic seeds keep this from reaching classical significance (Section \ref{sec:instability-replay-specialization}); on Lift, the benefit is not simply faster convergence. Over the full 3000-iteration run, most stuck baseline seeds never recover while most equally-stuck curriculum seeds do (Section \ref{sec:lift-extended-training}), yielding a 62.7-point IQM increase and reaching statistical significance.
This generalization extends to Factory, whose already-saturated baseline leaves little room to improve: the curriculum still raises IQM success rate by 2.2 points, reaching significance at this study's smallest sample size.
Across all three environments, the curriculum's impact is consistent: it reliably recovers seeds that would otherwise collapse, concentrates experience where the agent is currently struggling instead of scattering it across already-solved stages, and delivers higher final performance as a result.
The curriculum requires additional lightweight infrastructure and manual sub-task decomposition, but does not alter the environment, reward, or learning algorithm: evidence that reliable physical decision-making can be improved simply by rethinking where an agent's experience goes, without touching how it learns. Extending it to less structured domains such as locomotion and benchmarking against alternative curriculums are natural next steps.

\begin{ack}
This work received no external funding. The author declares no competing interests.
\end{ack}

\clearpage
\bibliographystyle{unsrtnat}
\bibliography{references}

@book{sutton1998reinforcement,
  title={Reinforcement learning: An introduction},
  author={Sutton, Richard S. and Barto, Andrew G.},
  volume={1},
  number={1},
  year={1998},
  publisher={MIT press Cambridge}
}

@article{mittal2025isaaclab,
  title={Isaac Lab: A GPU-Accelerated Simulation Framework for Multi-Modal Robot Learning},
  author={Mittal, Mayank and Roth, Pascal and Tigue, James and Richard, Antoine and Zhang, Octi and Du, Peter and Serrano-Mu{\~n}oz, Antonio and Yao, Xinjie and Zurbr{\"u}gg, Ren{\'e} and Rudin, Nikita and others},
  year={2025},
  journal={arXiv preprint arXiv:2511.04831}
}

@article{schwarke2025rslrl,
  title={RSL-RL: A Learning Library for Robotics Research},
  author={Schwarke, Clemens and Mittal, Mayank and Rudin, Nikita and Hoeller, David and Hutter, Marco},
  journal={arXiv preprint arXiv:2509.10771},
  year={2025}
}

@article{barto1983neuronlike,
  title={Neuronlike adaptive elements that can solve difficult learning control problems},
  author={Barto, Andrew G. and Sutton, Richard S. and Anderson, Charles W.},
  journal={IEEE Transactions on Systems, Man, and Cybernetics},
  volume={SMC-13},
  number={5},
  pages={834--846},
  year={1983}
}

@article{schulman2017proximal,
  title={Proximal policy optimization algorithms},
  author={Schulman, John and Wolski, Filip and Dhariwal, Prafulla and Radford, Alec and Klimov, Oleg},
  journal={arXiv preprint arXiv:1707.06347},
  year={2017}
}

@inproceedings{bengio2009curriculum,
  title={Curriculum learning},
  author={Bengio, Yoshua and Louradour, J{\'e}r{\^o}me and Collobert, Ronan and Weston, Jason},
  booktitle={Proceedings of the 26th annual international conference on machine learning},
  pages={41--48},
  year={2009}
}

@article{kumar2010self,
  title={Self-paced learning for latent variable models},
  author={Kumar, M and Packer, Benjamin and Koller, Daphne},
  journal={Advances in neural information processing systems},
  volume={23},
  year={2010}
}

@inproceedings{graves2017automated,
  title={Automated curriculum learning for neural networks},
  author={Graves, Alex and Bellemare, Marc G. and Menick, Jacob and Munos, Remi and Kavukcuoglu, Koray},
  booktitle={International conference on machine learning},
  pages={1311--1320},
  year={2017},
  organization={PMLR}
}

@article{matiisen2019teacher,
  title={Teacher--student curriculum learning},
  author={Matiisen, Tambet and Oliver, Avital and Cohen, Taco and Schulman, John},
  journal={IEEE transactions on neural networks and learning systems},
  volume={31},
  number={9},
  pages={3732--3740},
  year={2019},
  publisher={IEEE}
}

@inproceedings{florensa2017reverse,
  title={Reverse curriculum generation for reinforcement learning},
  author={Florensa, Carlos and Held, David and Wulfmeier, Markus and Zhang, Michael and Abbeel, Pieter},
  booktitle={Conference on robot learning},
  pages={482--495},
  year={2017},
  organization={PMLR}
}

@inproceedings{tobin2017domain,
  title={Domain randomization for transferring deep neural networks from simulation to the real world},
  author={Tobin, Josh and Fong, Rachel and Ray, Alex and Schneider, Jonas and Zaremba, Wojciech and Abbeel, Pieter},
  booktitle={2017 IEEE/RSJ International Conference on Intelligent Robots and Systems (IROS)},
  pages={23--30},
  year={2017},
  organization={IEEE}
}

@article{dennis2020emergent,
  title={Emergent complexity and zero-shot transfer via unsupervised environment design},
  author={Dennis, Michael and Jaques, Natasha and Vinitsky, Eugene and Bayen, Alexandre and Russell, Stuart and Critch, Andrew and Levine, Sergey},
  journal={Advances in neural information processing systems},
  volume={33},
  pages={13049--13061},
  year={2020}
}

@article{wang2019paired,
  title={Paired open-ended trailblazer (poet): Endlessly generating increasingly complex and diverse learning environments and their solutions},
  author={Wang, Rui and Lehman, Joel and Clune, Jeff and Stanley, Kenneth O.},
  journal={arXiv preprint arXiv:1901.01753},
  year={2019}
}

@article{ecoffet2019go,
  title={Go-explore: a new approach for hard-exploration problems},
  author={Ecoffet, Adrien and Huizinga, Joost and Lehman, Joel and Stanley, Kenneth O. and Clune, Jeff},
  journal={arXiv preprint arXiv:1901.10995},
  year={2019}
}

@article{zhou2024reset,
  title={Reset-Free Reinforcement Learning via Multi-State Recovery and Failure Prevention for Autonomous Robots},
  author={Zhou, Xu and Xu, Benlian and Jiang, Zhengqiang and Li, Jun and Nener, Brett},
  journal={Tsinghua Science and Technology},
  volume={29},
  number={5},
  pages={1481--1494},
  year={2024},
  publisher={TUP}
}

@inproceedings{henderson2018deep,
  title={Deep reinforcement learning that matters},
  author={Henderson, Peter and Islam, Riashat and Bachman, Philip and Pineau, Joelle and Precup, Doina and Meger, David},
  booktitle={Proceedings of the AAAI conference on artificial intelligence},
  volume={32},
  number={1},
  year={2018}
}

@inproceedings{de2018multi,
  title={Multi-step reinforcement learning: A unifying algorithm},
  author={De Asis, Kristopher and Hernandez-Garcia, J. Fernando and Holland, G. Zacharias and Sutton, Richard},
  booktitle={Proceedings of the AAAI conference on artificial intelligence},
  volume={32},
  number={1},
  year={2018}
}

@article{agarwal2021deep,
  title={Deep reinforcement learning at the edge of the statistical precipice},
  author={Agarwal, Rishabh and Schwarzer, Max and Castro, Pablo Samuel and Courville, Aaron C. and Bellemare, Marc},
  journal={Advances in neural information processing systems},
  volume={34},
  pages={29304--29320},
  year={2021}
}

\appendix

\section*{Appendix}
\label{sec:s0-appendix-intro}
This appendix contains material that supports the main paper's results and discussion. Each section below states which main-paper claim or section it supports. Nothing described in the main paper depends on this appendix to be understood; it exists to give an interested reader the full detail behind it.

\section{Full ablation study}
\label{sec:s1-ablations}

The main paper's Ablation Studies section covers three central ablations in depth (Confident Sampling, Exploratory Sampling, and Delay): Confident vs.\ Exploratory directly justifies why ACES blends the two strategies rather than picking one, and Delay is discussed in Section~\ref{sec:delayed-onset}. This section provides the full learning curves, prose, and statistics for the other eight ablations, plus consolidated statistics tables for all eleven.

\subsection{No EMA}
This ablation shows the effect of removing the EMA smoothing applied to success rates. An exponential moving average is normally used to smooth the success rates that feed the distribution calculation, reducing the batch-to-batch variability success rates otherwise exhibit. The EMA normally uses a coefficient $\alpha$ of 0.05; here, we report the result of removing this smoothing entirely.

\begin{figure}[H]
\centering
\includegraphics[width=\textwidth]{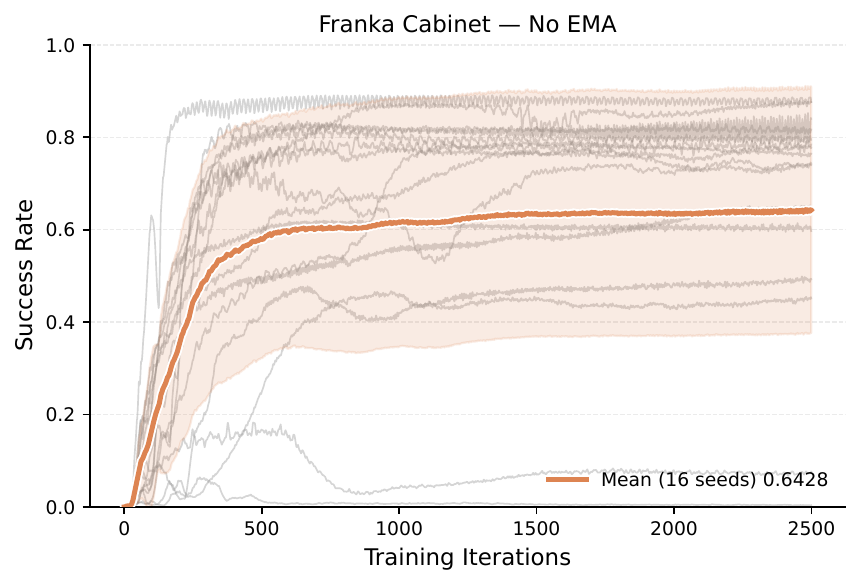}
\caption{Mean success rate over training with EMA smoothing of the success-rate signal removed, Franka Cabinet. Thin gray lines show individual seeds ($n=16$); shaded band shows $\pm 1$ sample standard deviation, clipped to $[0,1]$; the bold line shows the mean.}
\label{fig:s-ablation-no-ema-curve}
\end{figure}
Removing the EMA lowers the mean success rate relative to the main curriculum (0.6428 vs.\ 0.7032) and increases variance: at 0.2664 it is the highest sample standard deviation of any Cabinet condition reported anywhere in this paper, higher even than the main curriculum's own 0.2269. This is consistent with the main paper's Difficulty Estimation design (Section \ref{sec:difficulty-estimation}), where EMA smoothing was added specifically because noisy, unsmoothed success rates feed directly into curriculum instability.

\subsection{Curriculum sampling ratio 0.1 and 0.5}
These ablations report the results of lowering the curriculum sampling ratio to 10\%, and separately raising it to 50\%, relative to the typical 30\%, testing how much curriculum influence over resets the training process tolerates well.

\begin{figure}[H]
\centering
\begin{subfigure}[t]{0.48\textwidth}
\centering
\includegraphics[width=\textwidth]{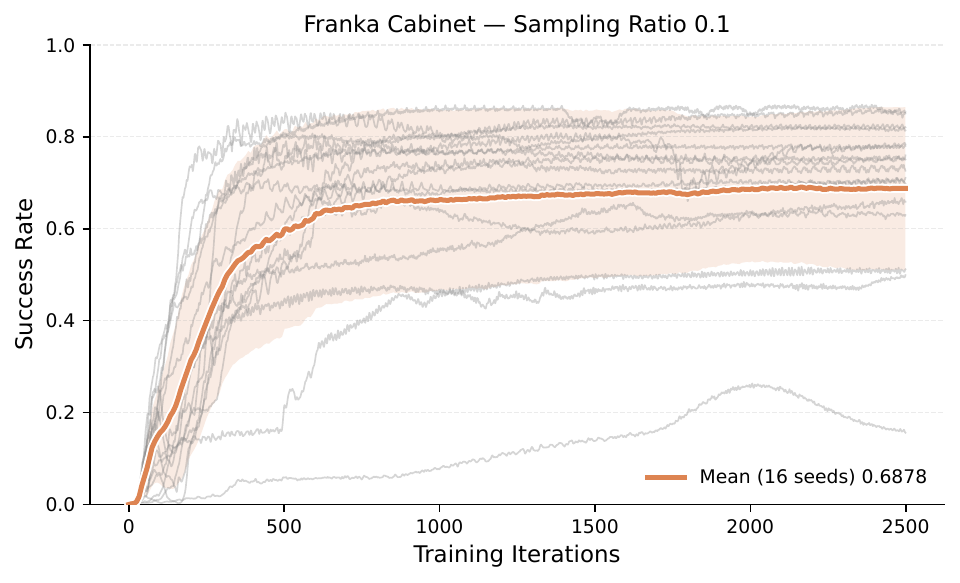}
\caption{Sampling ratio 0.1}
\label{fig:s-ablation-ratio-0p1-curve}
\end{subfigure}\hfill
\begin{subfigure}[t]{0.48\textwidth}
\centering
\includegraphics[width=\textwidth]{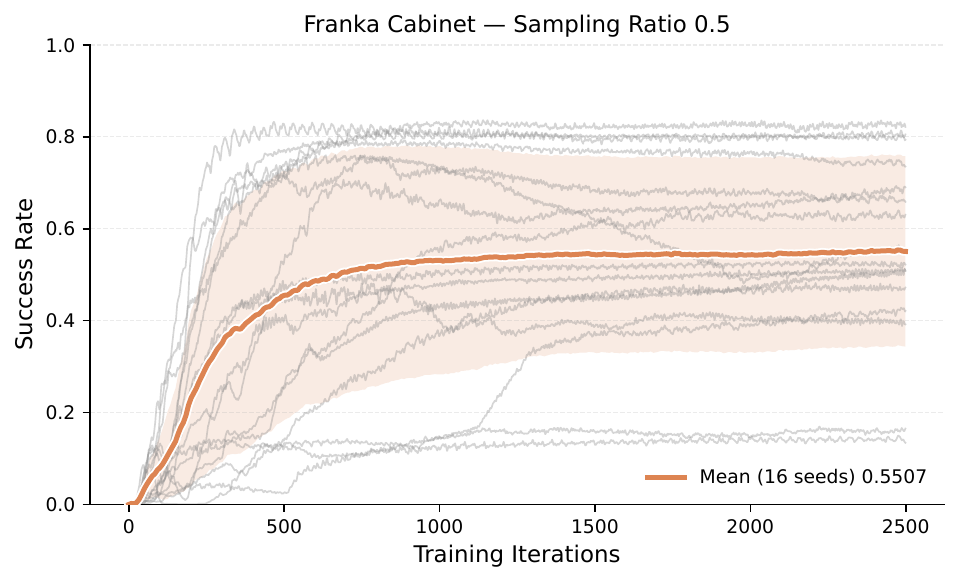}
\caption{Sampling ratio 0.5}
\label{fig:s-ablation-ratio-0p5-curve}
\end{subfigure}
\caption{Mean success rate over training with the curriculum sampling ratio lowered to 0.1 (a) and raised to 0.5 (b), Franka Cabinet. Thin gray lines show individual seeds ($n=16$); shaded band shows $\pm 1$ sample standard deviation, clipped to $[0,1]$; the bold line shows the mean.}
\label{fig:s-ablation-ratio-curves}
\end{figure}
Ratio 0.1 (mean 0.6878, median 0.7385, std 0.1770) tracks close to the main curriculum without reaching significance either way. This is expected as the curriculum has less effect as the ratio tends to zero, converting it back into a baseline training. Ratio 0.5 (mean 0.5507, median 0.5367, std 0.2071) is the one ablation in this entire study that reaches statistical significance in the \emph{unfavorable} direction against baseline (95\% CI $[0.063, 0.438]$, Cliff's $\delta=-0.50$, Wilcoxon $p=0.044$): direct evidence that over-weighting curriculum-driven resets actively hurts training rather than merely failing to help, consistent with the main paper's finding that pushing the sampling ratio to 0.5 degrades performance sharply (Section \ref{sec:instability-replay-specialization}, Curriculum Hyperparameter Interactions).

\subsection{Buffer size 32 and 128}
This ablation alters the success buffer size from the default 64 to 32 and to 128, directly changing how the curriculum stores world states as sub-tasks are completed.

\begin{figure}[H]
\centering
\begin{subfigure}[t]{0.48\textwidth}
\centering
\includegraphics[width=\textwidth]{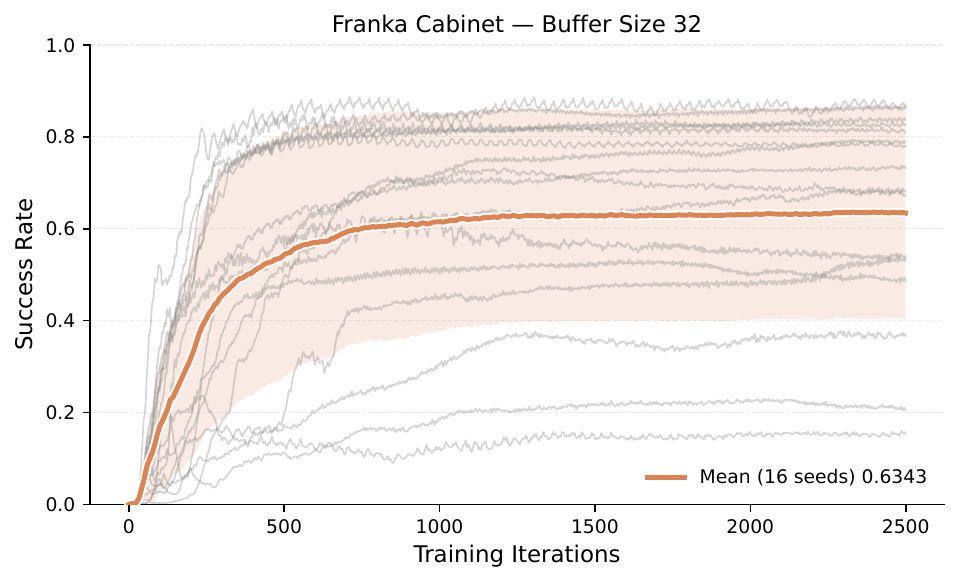}
\caption{Buffer size 32}
\label{fig:s-ablation-buffer-32-curve}
\end{subfigure}\hfill
\begin{subfigure}[t]{0.48\textwidth}
\centering
\includegraphics[width=\textwidth]{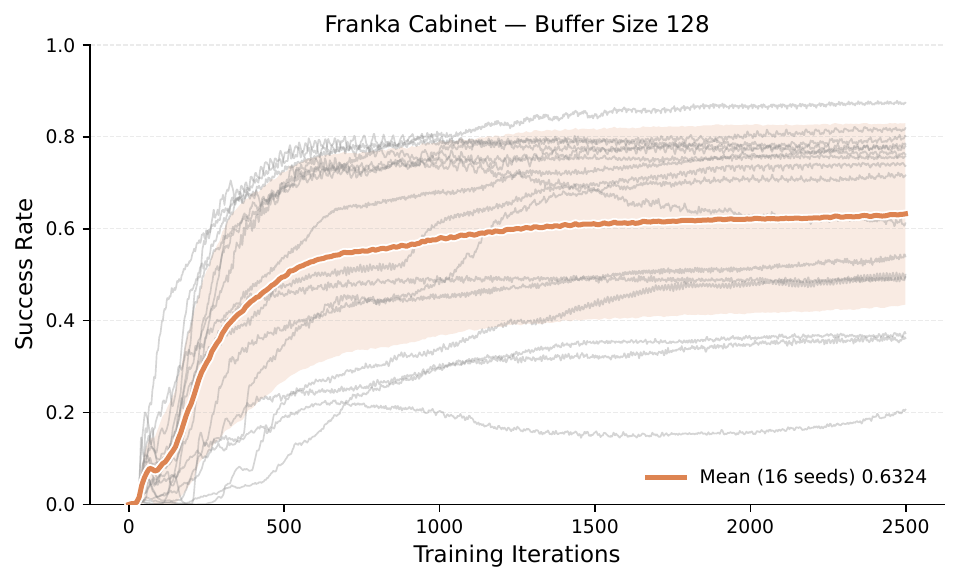}
\caption{Buffer size 128}
\label{fig:s-ablation-buffer-128-curve}
\end{subfigure}
\caption{Mean success rate over training with the success buffer size reduced to 32 (a) and increased to 128 (b), Franka Cabinet. Thin gray lines show individual seeds ($n=16$); shaded band shows $\pm 1$ sample standard deviation, clipped to $[0,1]$; the bold line shows the mean.}
\label{fig:s-ablation-buffer-curves}
\end{figure}
Buffer 32 (mean 0.6343, median 0.7067, std 0.2307) and Buffer 128 (mean 0.6324, median 0.7266, std 0.1974) both trail the main curriculum without reaching individual significance, though Buffer 128 shows a meaningful negative directional shift (95\% CI $[0.125, 0.563]$ against baseline).

\subsection{Velocity}
This ablation adds the joint velocities of the robot arm and the cabinet into the world representation, carrying additional data into the reset.

\begin{figure}[H]
\centering
\includegraphics[width=\textwidth]{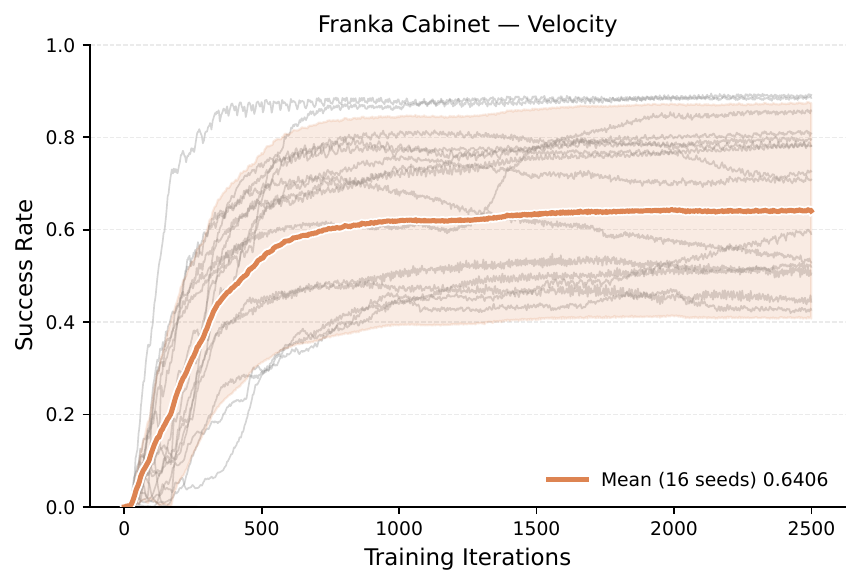}
\caption{Mean success rate over training with joint velocities included in the world representation, Franka Cabinet. Thin gray lines show individual seeds ($n=16$); shaded band shows $\pm 1$ sample standard deviation, clipped to $[0,1]$; the bold line shows the mean.}
\label{fig:s-ablation-velocity-curve}
\end{figure}
Including velocities lowers both mean (0.6406 vs.\ 0.7032) and median (0.7173 vs.\ 0.8073) success rate relative to the main curriculum, without any corresponding reduction in variance (std 0.2333 vs.\ 0.2269). This supports the hypothesis (Section \ref{sec:design-guidelines}) that replaying stored velocities as though the agent had just caused that motion introduces a confusing, uncaused signal rather than useful information. Adding more information to the world representation is not always helpful for the agent's learning.

\subsection{Pure sub-task replay (default to sub-task 3)}
\label{sec:s-pure-subtask-replay}
This ablation ignores the distribution algorithm entirely and always targets sub-task 3 for every curriculum episode, meaning it continuously resets to the point where the agent is touching the drawer handle but has not yet pulled it open. It isolates whether the curriculum's benefit comes from \emph{targeted} sampling of the agent's actual bottleneck, or whether simply giving the agent more general replay practice, at any fixed sub-task, regardless of where it is currently struggling, is what drives the improvement.

\begin{figure}[H]
\centering
\includegraphics[width=\textwidth]{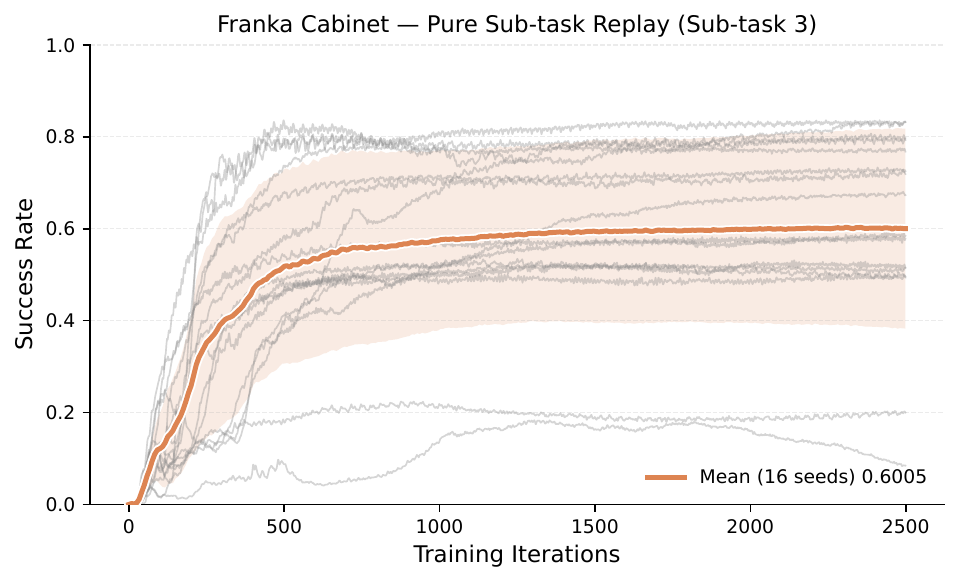}
\caption{Mean success rate over training with the curriculum always targeting sub-task 3 (Pure Sub-task Replay), Franka Cabinet. Thin gray lines show individual seeds ($n=16$); shaded band shows $\pm1$ sample standard deviation, clipped to $[0,1]$; bold line shows the mean.}
\label{fig:s-ablation-default-3-curve}
\end{figure}
Always replaying to sub-task 3 (mean 0.6005, median 0.6289, std 0.2176, IQM 0.6348) underperforms both the full ACES blend (mean 0.7032, IQM 0.7995) and the no-curriculum baseline (mean 0.7004, IQM 0.7072) on every descriptive statistic. The comparison against baseline shows a meaningful negative directional shift (95\% CI $[0.125, 0.563]$, Cliff's $\delta=-0.375$) that falls just short of classical significance (Wilcoxon $p=0.0934$, Table \ref{tab:s-ablation-significance}); against ACES the difference is smaller and not directionally meaningful under the paper's own threshold (95\% CI $[0.188, 0.688]$).

Originally we hypothesized that fixed-sub-task replay alone might offer a modest, curriculum-independent benefit. The full 16-seed result instead shows fixed-sub-task replay is, if anything, directionally worse than doing nothing at all (i.e.\ the no-curriculum baseline), and does not reproduce ACES's improvement. It supports the interpretation that the curriculum's benefit comes specifically from targeting the agent's actual, currently-estimated bottleneck (Section \ref{sec:distribution-algorithms} of the main paper), not merely from spending more training time inside sub-task replay in general. Naive fixed-target replay appears to invite the same narrow, non-generalizing specialization risk as targeted replay, without the compensating benefit of resolving the agent's real bottleneck.
Replay alone is not the source of the curriculum's benefit: empirically-driven bottleneck selection is. Letting the agent's own recorded, per-sub-task performance decide \emph{where} training resumes is what helps, not simply giving it more resets at a fixed, hand-picked point.

\subsection{Eval isolation (curriculum/non-curriculum separation removed)}
\label{sec:s-eval-isolation}
This ablation removes the separation between curriculum and non-curriculum episodes when computing the success-rate signal that difficulty estimation depends on (main paper, Overview and Difficulty Estimation subsections). Every episode, whether or not its reset pose was chosen by the curriculum, now contributes to the measured success rate, reintroducing the eval-contamination risk the main design explicitly isolates against.

\begin{figure}[H]
\centering
\includegraphics[width=\textwidth]{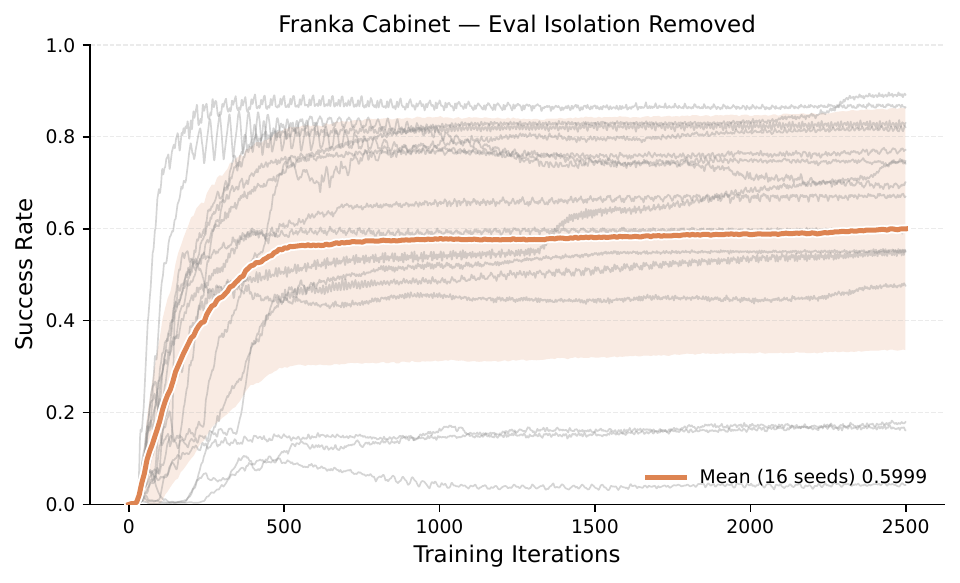}
\caption{Mean success rate over training with curriculum/non-curriculum episode separation removed from the success-rate signal (Eval Isolation Removed), Franka Cabinet. Thin gray lines show individual seeds ($n=16$); shaded band shows $\pm1$ sample standard deviation, clipped to $[0,1]$; bold line shows the mean.}
\label{fig:s-ablation-no-isolation-curve}
\end{figure}
Removing eval isolation (mean 0.5999, median 0.6856, std 0.2637, IQM 0.6665) also underperforms ACES on every descriptive statistic, and is among the noisiest conditions in this study by standard deviation (0.2637, second only to No EMA's 0.2664). This condition shows a meaningful negative directional shift against both baseline (95\% CI $[0.125, 0.625]$, $\delta=-0.25$) and ACES (95\% CI $[0.125, 0.625]$, $\delta=-0.25$); the comparison against the baseline does not reach significance ($p=0.3755$), while comparison with ACES approaches but does not reach significance (Wilcoxon $p=0.0934$).

This is consistent with the main paper's core methodological argument for isolating the success-rate signal in the first place: the eval-contamination check shown in Figure~\ref{fig:replay-vs-eval} already showed curriculum episodes reaching a substantially higher success rate (0.9215) than genuine non-curriculum performance (0.7407), direct evidence the two populations differ enough that mixing them would bias the signal. This ablation shows what happens when that bias is actually allowed into the difficulty estimate: performance degrades relative to both the isolated curriculum and, directionally, relative to no curriculum at all.
\subsection{Overall performance: descriptive statistics}
The main paper's Results section (Table \ref{tab:math-analysis}) reports IQM and paired-comparison statistics for the headline Cabinet, Lift, and Factory comparisons; full descriptive statistics (mean, median, standard deviation, min, max) for those same conditions are collected here.

\begin{table}[H]
\centering
\caption{Overall performance: descriptive statistics per condition (final success rate). Cabinet and Lift use $n=16$ seeds; Factory uses $n=8$ (Section \ref{sec:baselines}), with each seed's value taken as its peak over the final 25 logged steps (Figure~\ref{fig:three-environments}'s caption, Appendix~\ref{sec:s3-hyperparams}).}
\label{tab:overall-performance}
\small
\renewcommand{\arraystretch}{1.2}
\begin{tabular}{lccccc}
\toprule
\textbf{Condition} & \textbf{Mean} & \textbf{Median} & \textbf{Std.} & \textbf{Min} & \textbf{Max} \\
\midrule
Cabinet Baseline & 0.7004 & 0.7321 & 0.1137 & 0.5500 & 0.8345 \\
Cabinet Experimental (ACES) & 0.7032 & 0.8073 & 0.2269 & 0.1088 & 0.8909 \\
Factory Baseline & 0.9102 & 0.9110 & 0.0097 & 0.8974 & 0.9255 \\
Factory Experimental (ACES) & 0.9308 & 0.9327 & 0.0119 & 0.9094 & 0.9458 \\
Lift Baseline & 0.4310 & 0.1222 & 0.4741 & 0.0000 & 1.0000 \\
Lift Experimental (ACES) & 0.8535 & 0.9957 & 0.3366 & 0.0000 & 1.0000 \\
\bottomrule
\end{tabular}
\renewcommand{\arraystretch}{1.0}
\end{table}

\subsection{Full ablation statistics}

\begin{table}[H]
\centering
\caption{Descriptive statistics for all 11 trained ablations plus the main curriculum and baseline, Franka Cabinet, $n=16$ seeds each. IQM for each condition is reported alongside its significance statistics in Table \ref{tab:s-ablation-significance}.}
\label{tab:s-ablation-descriptive}
\small
\begin{tabular}{lccccc}
\toprule
\textbf{Condition} & \textbf{Mean} & \textbf{Median} & \textbf{Std.} & \textbf{Min} & \textbf{Max} \\
\midrule
Baseline & 0.7004 & 0.7321 & 0.1137 & 0.5500 & 0.8345 \\
Main Curriculum (ACES) & 0.7032 & 0.8073 & 0.2269 & 0.1088 & 0.8909 \\
Sampling Ratio 0.1 & 0.6878 & 0.7385 & 0.1770 & 0.1561 & 0.8539 \\
Sampling Ratio 0.5 & 0.5507 & 0.5367 & 0.2071 & 0.1343 & 0.8215 \\
Buffer Size 32 & 0.6343 & 0.7067 & 0.2307 & 0.1547 & 0.8663 \\
Buffer Size 128 & 0.6324 & 0.7266 & 0.1974 & 0.2064 & 0.8741 \\
Confident (Argmax) & 0.6484 & 0.7369 & 0.2027 & 0.2218 & 0.8857 \\
Exploratory (Softmax) & 0.5721 & 0.6459 & 0.2601 & 0.0125 & 0.8488 \\
No EMA & 0.6428 & 0.7537 & 0.2664 & 0.0037 & 0.8765 \\
Delay & 0.7198 & 0.7692 & 0.1613 & 0.1953 & 0.8642 \\
Velocity & 0.6406 & 0.7172 & 0.2333 & 0.0000 & 0.8917 \\
Pure Sub-task Replay (Sub-task 3) & 0.6005 & 0.6289 & 0.2176 & 0.0847 & 0.8324 \\
Eval Isolation Removed & 0.5999 & 0.6856 & 0.2637 & 0.0424 & 0.8916 \\
\bottomrule
\end{tabular}
\end{table}

\begin{table}[H]
\centering
\caption{Paired comparison statistics, all 11 ablations vs.\ baseline and vs.\ the main curriculum (ACES), $n=16$ paired by seed. IQM is each ablation's own descriptive statistic (Table \ref{tab:s-ablation-descriptive}), repeated across both comparison blocks since it does not depend on which baseline it is compared against. Cliff's $\delta$ above $+0.20$ or below $-0.20$ (equivalently, $P(X{>}Y)$ above 0.60 or below 0.40) is flagged as a meaningful directional shift per the main paper's stated threshold, independent of classical significance. 95\% CI is the bootstrap CI on $P(X{>}Y)$ ($B=10{,}000$ paired resamples). \textbf{Bold} Wilcoxon $p$ denotes statistical significance ($p<0.05$).}
\label{tab:s-ablation-significance}
\small
\begin{tabular}{lccccc}
\toprule
\textbf{Comparison} & \textbf{IQM} & \textbf{95\% CI} & \textbf{Cliff's $\delta$} & \textbf{Wilcoxon $p$} \\
\midrule
\multicolumn{5}{l}{\textit{vs.\ Baseline}} \\
Sampling Ratio 0.1 & 0.7335 & $[0.313, 0.813]$ & +0.125 & 0.4954 \\
Sampling Ratio 0.5 & 0.5677 & $[0.063, 0.438]$ & --0.500 & \textbf{0.0443} \\
Buffer Size 32 & 0.6923 & $[0.250, 0.750]$ & +0.000 & 0.4637 \\
Buffer Size 128 & 0.6755 & $[0.125, 0.563]$ & --0.375 & 0.1754 \\
Confident (Argmax) & 0.6892 & $[0.188, 0.688]$ & --0.125 & 0.5282 \\
Exploratory (Softmax) & 0.6272 & $[0.250, 0.750]$ & +0.000 & 0.3225 \\
No EMA & 0.7314 & $[0.313, 0.813]$ & +0.125 & 1.0000 \\
Delay & 0.7647 & $[0.313, 0.813]$ & +0.125 & 0.4037 \\
Velocity & 0.6790 & $[0.125, 0.563]$ & --0.375 & 0.3484 \\
Pure Sub-task Replay (Sub-task 3) & 0.6348 & $[0.125, 0.563]$ & --0.375 & 0.0934 \\
Eval Isolation Removed & 0.6665 & $[0.125, 0.625]$ & --0.250 & 0.3755 \\
\midrule
\multicolumn{5}{l}{\textit{vs.\ Main Curriculum (ACES)}} \\
Sampling Ratio 0.1 & 0.7335 & $[0.125, 0.563]$ & --0.375 & 0.4637 \\
Sampling Ratio 0.5 & 0.5677 & $[0.063, 0.500]$ & --0.500 & 0.0739 \\
Buffer Size 32 & 0.6923 & $[0.125, 0.563]$ & --0.375 & 0.2744 \\
Buffer Size 128 & 0.6755 & $[0.000, 0.375]$ & --0.625 & 0.1167 \\
Confident (Argmax) & 0.6892 & $[0.063, 0.500]$ & --0.500 & 0.0739 \\
Exploratory (Softmax) & 0.6272 & $[0.125, 0.563]$ & --0.375 & 0.1591 \\
No EMA & 0.7314 & $[0.125, 0.625]$ & --0.250 & 0.1439 \\
Delay & 0.7647 & $[0.125, 0.563]$ & --0.375 & 0.5282 \\
Velocity & 0.6790 & $[0.125, 0.563]$ & --0.375 & 0.3484 \\
Pure Sub-task Replay (Sub-task 3) & 0.6348 & $[0.188, 0.688]$ & --0.125 & 0.2522 \\
Eval Isolation Removed & 0.6665 & $[0.125, 0.625]$ & --0.250 & 0.0934 \\
\bottomrule
\end{tabular}
\end{table}

Ten of the eleven ablations are directionally beaten by the full ACES blend (Cliff's $\delta < -0.20$ against ACES), though none individually reaches significance in that comparison at $n=16$. This is consistent with the main paper's framing that no single component swap matches the full method, without overclaiming any one ablation as individually disproven.

\subsection{Performance profiles}
\begin{figure}[H]
\centering
\includegraphics[width=\textwidth]{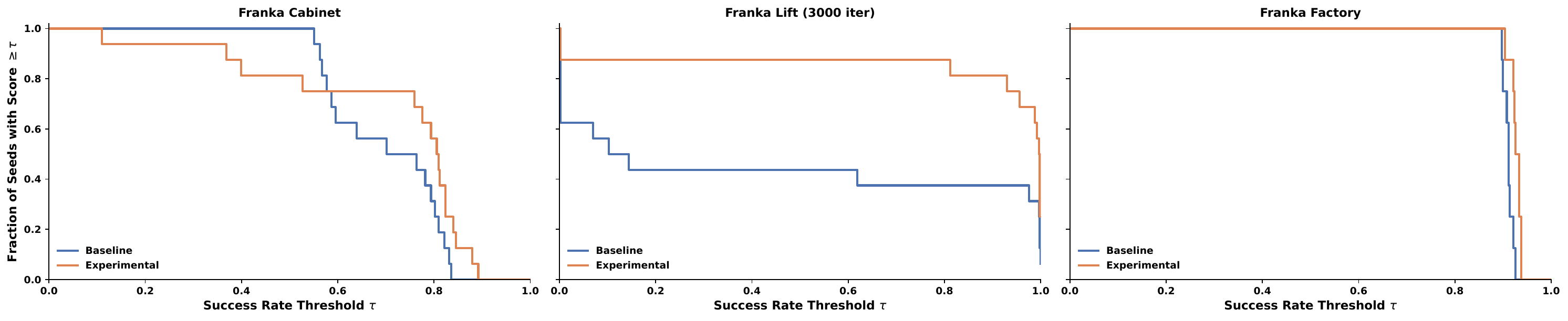}
\caption{Performance profiles: for each threshold $\tau$ on the x-axis, the y-axis gives the fraction of seeds whose final success rate is at least $\tau$, baseline (blue) vs.\ main curriculum (orange), Cabinet (left, $n=16$), Lift (center, $n=16$), and Factory (right, $n=8$).}
\label{fig:s-performance-profiles}
\end{figure}
Lift's curriculum dominates the baseline at every threshold; Cabinet's curriculum raises the ceiling ($31\%\to56\%$ of seeds reach $\geq 0.8$) but lowers the floor (its worst seed sits below the baseline's worst), the quantitative basis for the main paper's replay-specialization argument (Section \ref{sec:instability-replay-specialization}). Factory's curriculum profile sits uniformly above baseline despite the narrow, near-ceiling range both conditions occupy, the same saturated-baseline pattern discussed in Section~\ref{sec:results}.

\section{Well-performing vs.\ catastrophic seed diagnostic}
\label{sec:s-good-bad-seed}

The main paper's Replay Specialization argument (Section \ref{sec:instability-replay-specialization}, ``Diagnostic Evidence for Replay Specialization'') is supported in prose by a comparison of reset-pose deviation, entropy, and loss between well-performing and catastrophic Cabinet seeds. This figure makes that comparison visible directly and at greater statistical weight than a single pair: the top 4 curriculum seeds by final success rate (54, 42, 50, 55; success rates 0.891--0.838) against the bottom 4 (48, 46, 47, 43; success rates 0.109--0.527), all from the main Cabinet curriculum run.

\begin{figure}[H]
\centering
\includegraphics[width=\textwidth]{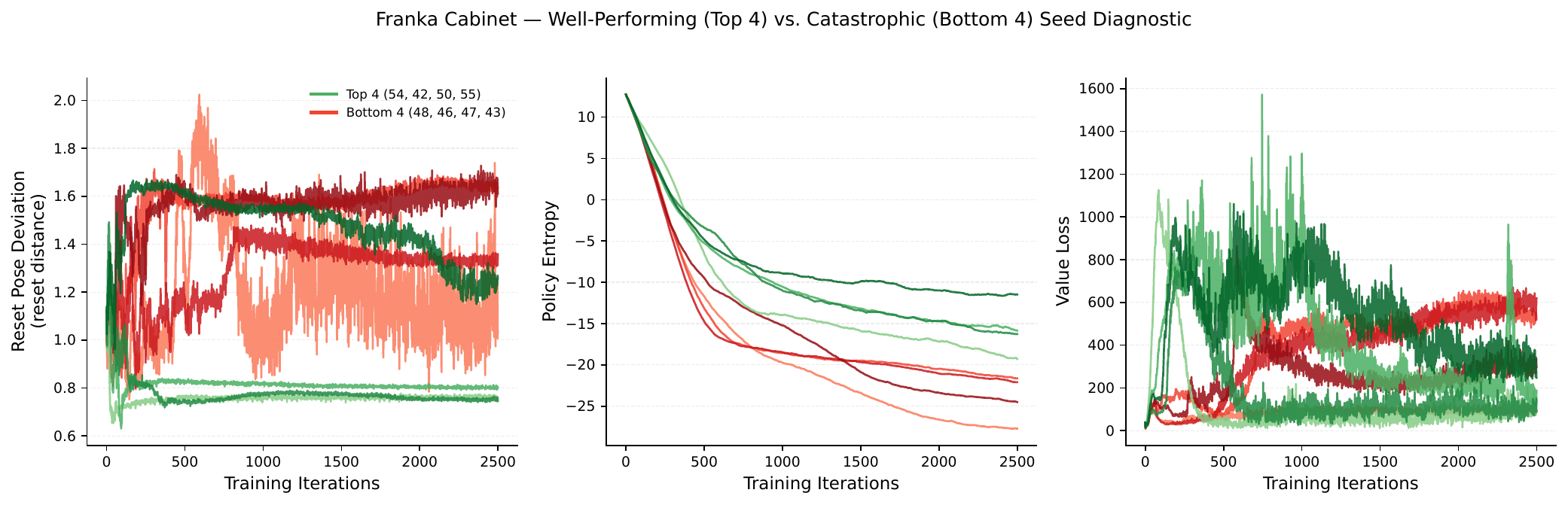}
\caption{Reset pose deviation (left), policy entropy (center), and value loss (right) over training, top-4 well-performing seeds (green shades) vs.\ bottom-4 catastrophic seeds (red shades), Franka Cabinet main curriculum run.}
\label{fig:s-good-vs-bad-seed}
\end{figure}

\textbf{Reset pose deviation}: the group means separate clearly (top-4 mean 0.886 vs.\ bottom-4 mean 1.429 over the final 10\% of training), but individual seeds show more spread than the entropy panel: three of the four top seeds (54, 42, 50) settle to a low, stable value ($\approx$0.75--0.80) while the fourth (55) sits closer to the bottom group's range ($\approx$1.23) as it slowly declines. The group-level pattern (well-performing seeds converge to a lower, more stable reset-pose deviation) holds, but it is not a uniform per-seed rule at $n=4$ per group.

\textbf{Entropy}: cleanly separated across the full 16-seed groups, not just a coincidence of one pair. The top-4 group averages $-15.42$ over the final 10\% of training vs.\ $-23.71$ for the bottom-4 group (per-seed range: top $-18.7$ to $-11.5$; bottom $-27.6$ to $-21.4$, no overlap). This matches the main paper's claim that catastrophic seeds' policies become confidently deterministic without that confidence translating to task success.

\textbf{Value loss}: at $n=8$ the pattern the main paper describes holds cleanly and is stated there as a confirmed finding, not an open discrepancy. All four top-4 seeds show a pronounced early loss spike (peak-to-final-value ratio 3.3--10.6$\times$, peaking within the first 30\% of training) that then declines substantially. The bottom-4 seeds show no comparable spike (peak-to-final ratio 1.1--2.9$\times$), and two of the four (46, 47) peak very late in training (87\% and 94\% of the way through) rather than early, consistent with a gradual increase rather than an early spike-and-decline. This loss behavior often signifies the agent's critic network cannot predict or conceptualize the environment.
\section{Standard RL training diagnostics}
\label{sec:s2-diagnostics}

The main paper's Baselines and Metrics section (Section~\ref{sec:baselines}) notes that a standard set of reinforcement-learning training metrics is tracked throughout every run: entropy, value loss, learning rate, surrogate loss, rewards, collection/learning time, mean policy standard deviation, and episode length. These are all logged to TensorBoard throughout training. The curriculum-specific diagnostics shown below (reset distance and variance, selected sub-task, difficulty, gap, margin, blend, sampling distribution, and realized sampling rate) are specific to this paper's curriculum and are not standard RL vocabulary; full definitions are in Appendix~\ref{sec:s5-keywords}'s glossary.

The two figures below give every reader a direct look at these values over training, rather than requiring them to be inferred from prose. Both are computed from the main Cabinet curriculum run, mean $\pm$ one standard deviation across all 16 seeds; each panel carries a one-line caption explaining what it measures and why it matters.

\begin{figure}[H]
\centering
\includegraphics[width=\textwidth]{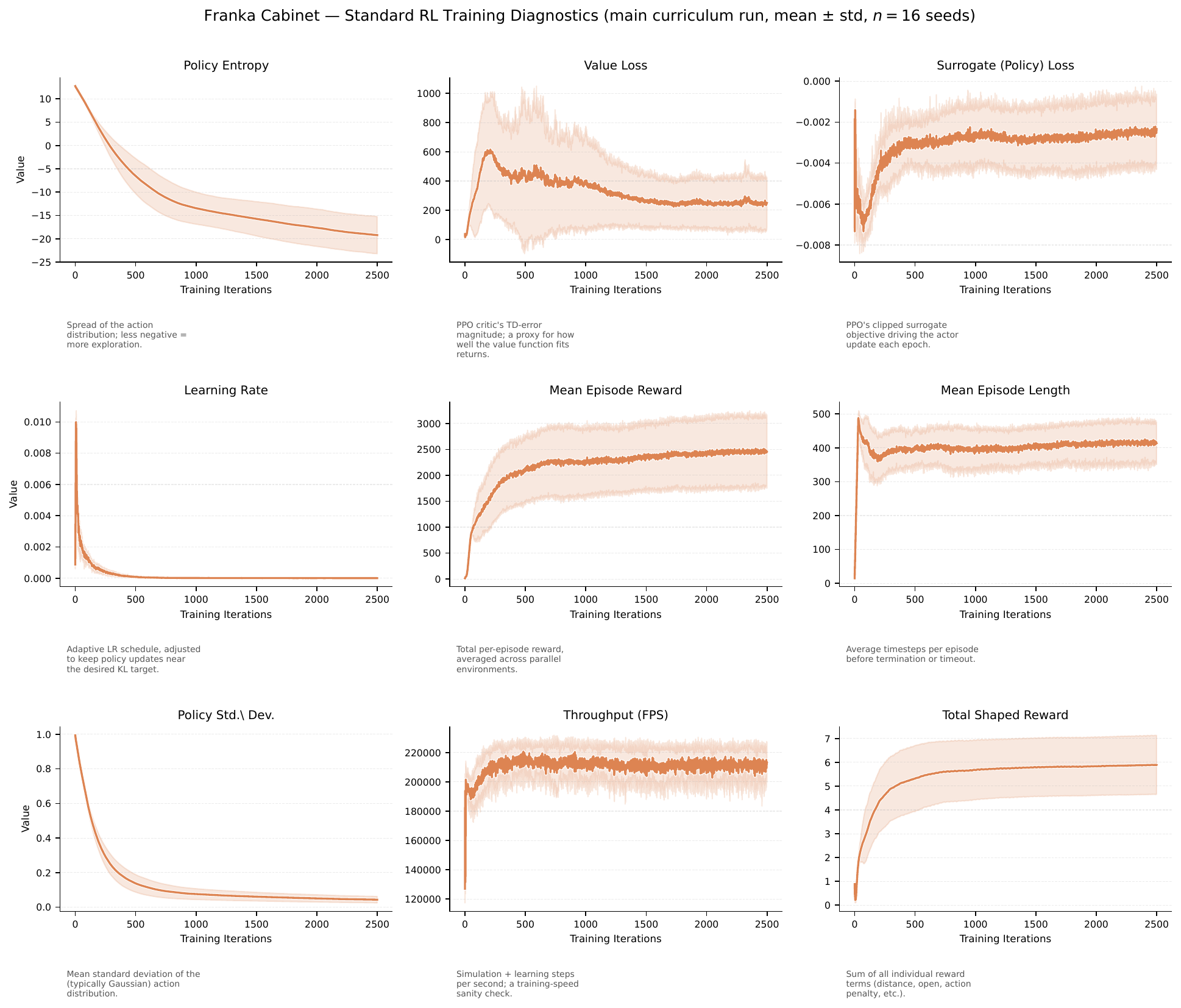}
\caption{Standard reinforcement-learning training diagnostics for the Franka Cabinet main curriculum run (mean $\pm$ std, $n=16$ seeds): policy entropy, value loss, surrogate (policy) loss, learning rate, mean episode reward, mean episode length, policy standard deviation, simulation throughput, and total shaped reward. Confirms PPO training behaves as expected (entropy and learning rate decay smoothly, reward and episode length converge) with no signs of the curriculum itself destabilizing the underlying RL algorithm.}
\label{fig:s-rl-diagnostics}
\end{figure}

\begin{figure}[H]
\centering
\includegraphics[width=\textwidth]{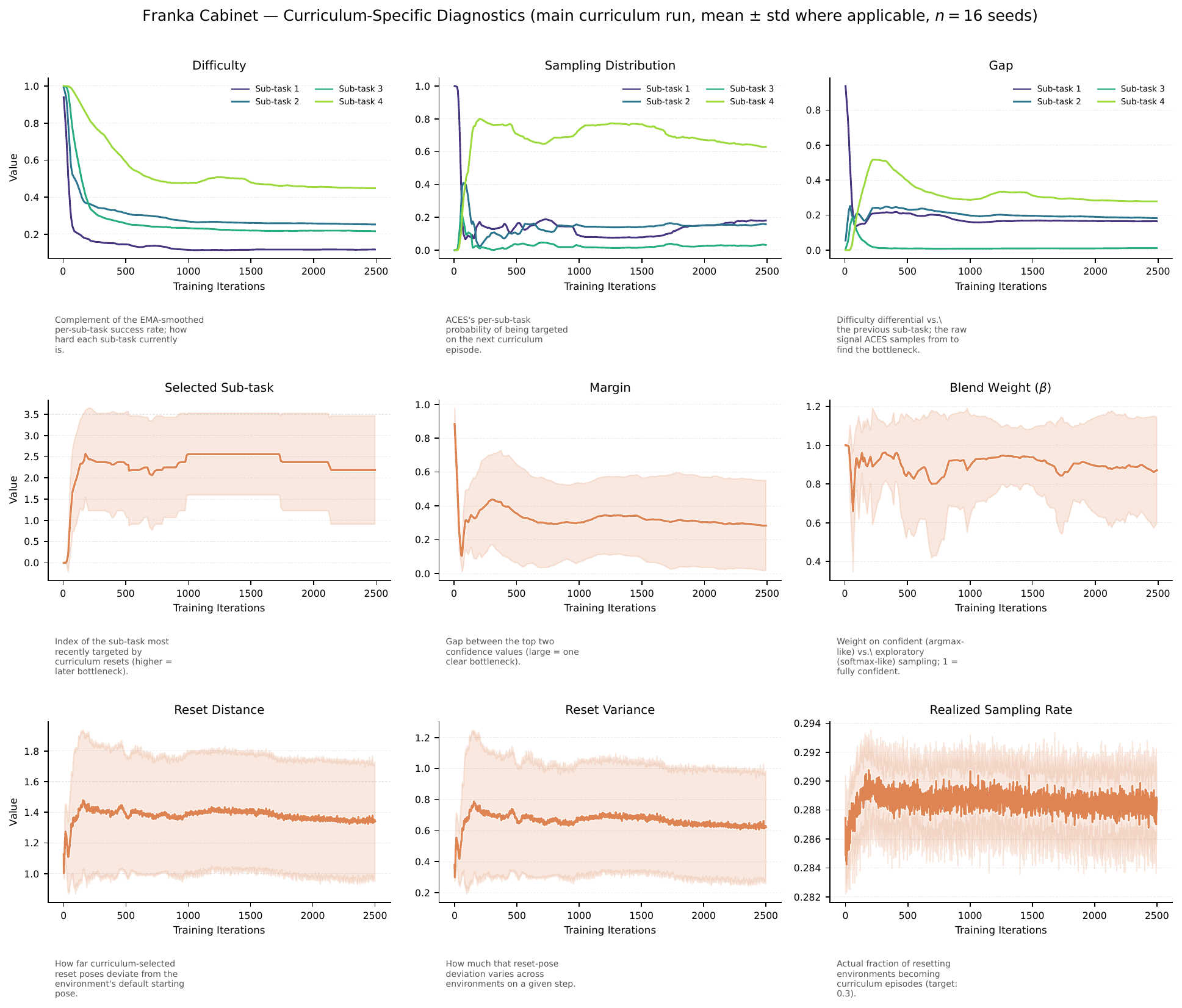}
\caption{Curriculum-specific diagnostics for the same run: per-sub-task difficulty, sampling distribution, and gap (each with 4 lines, one per sub-task); selected sub-task, margin, and blend weight; reset distance, reset variance, and realized sampling rate. Difficulty and gap together show the curriculum correctly tracking sub-task 4 as the persistent bottleneck (highest difficulty and, after an early peak, still the largest gap) while margin and blend weight show ACES settling into a mostly-confident ($\beta \approx 0.85$--$0.95$) sampling regime rather than oscillating between modes.}
\label{fig:s-curriculum-diagnostics}
\end{figure}

Two related diagnostics round out this appendix: one supports the main paper's eval-contamination argument (Section \ref{sec:overview}) by comparing curriculum-episode (``replay'') success rate against non-curriculum-episode (``eval'') success rate directly; the other shows ACES's blend weight adapting over training rather than settling into one fixed regime.

\begin{figure}[H]
\centering
\begin{subfigure}[t]{0.48\textwidth}
\centering
\includegraphics[width=\textwidth]{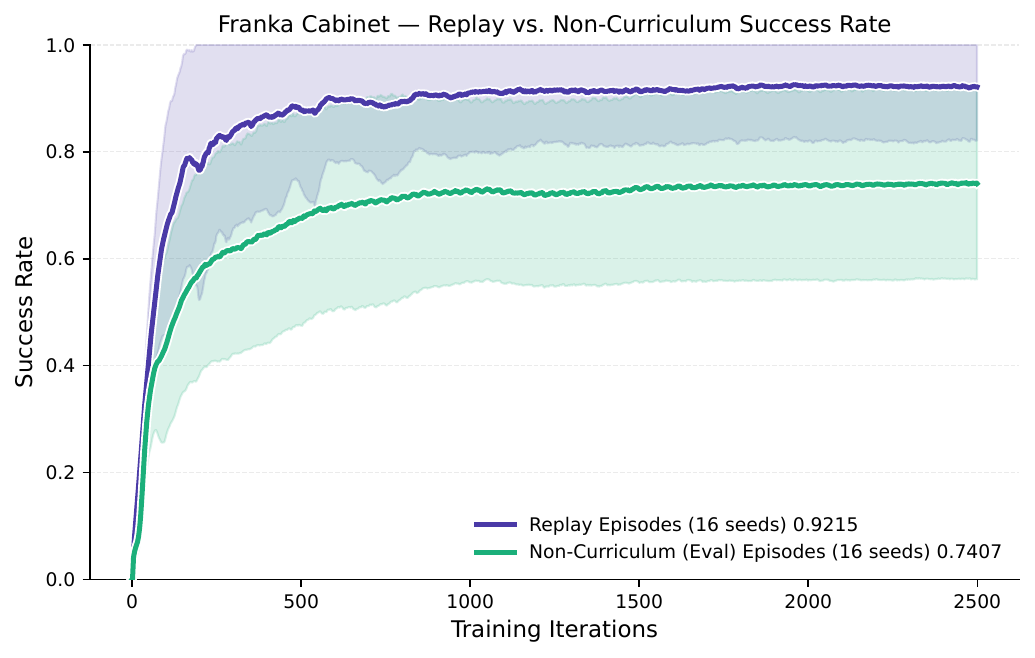}
\caption{Replay vs.\ eval success rate}
\label{fig:replay-vs-eval}
\end{subfigure}\hfill
\begin{subfigure}[t]{0.48\textwidth}
\centering
\includegraphics[width=\textwidth]{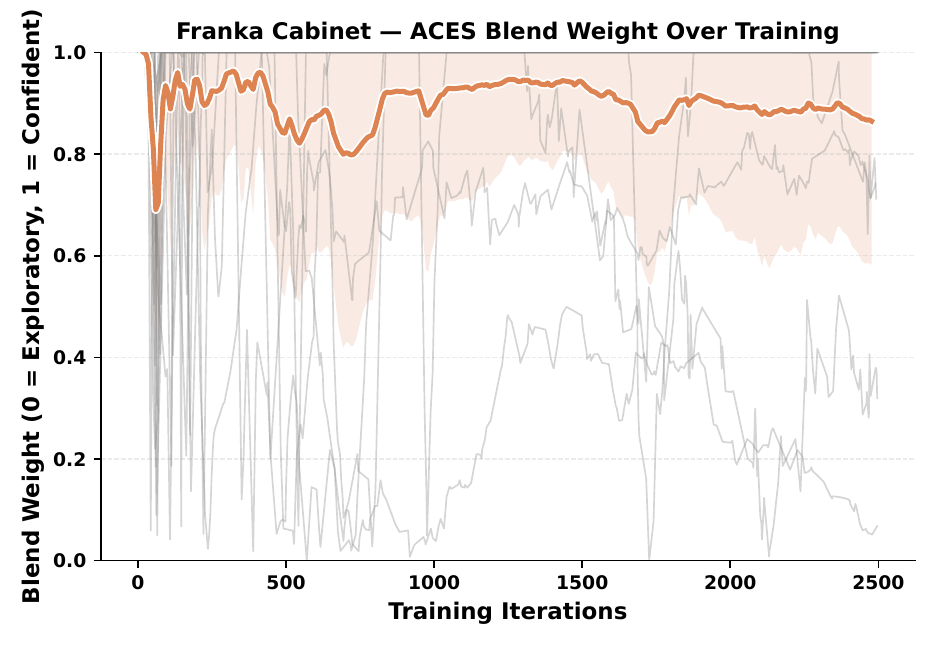}
\caption{ACES blend weight over training}
\label{fig:blend-weight-cabinet}
\end{subfigure}
\caption{(a) Mean success rate over training, Franka Cabinet, comparing curriculum-episode (``replay'') success rate against non-curriculum-episode (``eval'') success rate, mean across 16 seeds $\pm 1$ sample standard deviation. (b) ACES blend weight $\beta$ over training, Franka Cabinet, mean across 16 seeds (orange) $\pm 1$ sample standard deviation (shaded band), with individual seed traces in gray. $\beta=1$ is purely confident ($\operatorname{argmax}$) sampling, $\beta=0$ is purely exploratory ($\operatorname{softmax}$) sampling.}
\label{fig:diagnostics-eval-blend}
\end{figure}
Replay episodes reach a substantially higher success rate (0.9215) than genuine non-curriculum performance (0.7407), confirming that curriculum episodes would meaningfully bias the difficulty signal if not isolated from it (Section \ref{sec:results}). The individual gray seed traces in (b) show this is not just an aggregate effect: seed by seed, blend weight repeatedly spikes down toward 0 (exploratory sampling) at various points in training as that seed's own difficulty margin temporarily narrows, then partially recovers as its bottleneck resolves. The mean's dip to $\beta\approx0.70$ around iteration 0--200 (Section \ref{sec:distribution-algorithms}) reflects the dense sub-task difficulties found early in the training, which promotes more exploratory weighting accordingly.

\section{General hyperparameters and sub-task definitions}
\label{sec:s3-hyperparams}

The main paper's Experimental Setup section gives a condensed, prose description of each environment's sub-tasks and training configuration, pointing here for full detail.

\textbf{World state storage conventions.} Supports Section~\ref{sec:world-state-modeling}. Fixed objects (e.g.\ a cabinet body) omit orientation and position from their stored world state, since these never change; only moving parts (the robot arm, articulated joints, manipulated objects) store full pose. All stored positions are relative to the environment origin and are added back at reset time, so a single world state is valid regardless of which parallel environment it was recorded in or resets into. The creation of world representations should consider that additional information may have unintended consequences. We found that more quantities in the representation can have negative impacts, which you can see in Appendix~\ref{sec:s1-ablations}

\textbf{Software licenses.} All environments and training infrastructure build on existing open-source packages, used as-is and cited in Section~5.1: IsaacLab~\cite{mittal2025isaaclab} (BSD-3-Clause), RSL-RL~\cite{schwarke2025rslrl} (BSD-3-Clause), and RL-Games (MIT License).

\textbf{Compute resources.} All experiments were run on a single workstation: an NVIDIA RTX 5090 (32\,GB VRAM), an AMD Ryzen 9950X3D CPU, and 64\,GB system RAM.

\textbf{Factory success-rate extraction.} Supports Section~\ref{sec:results} and Figure~\ref{fig:three-environments}. Due to the way the Factory environment was set up, its per-step logged success rate oscillates with a period of roughly 3--4 iterations, occasionally up to 7: RL-Games logs a success flag for whichever parallel environments happen to have completed an episode as of that exact step, and since Factory's episodes span multiple iterations, most logged steps catch a window in which few or no environments have completed one, undersampling the true rate. The envelope traced by each cycle's peak is consistent across the final several cycles of training, indicating genuine convergence rather than a favorable sample. Reported Factory statistics therefore use each seed's maximum value over its final 25 logged steps (covering several oscillation periods) as the converged success rate, rather than the literal final logged value used for Cabinet and Lift, whose logging does not exhibit this artifact; plotted curves track a trailing 9-step rolling maximum per seed for the same reason.

\begin{table}[H]
\centering
\caption{Sub-task definitions and completion thresholds for each benchmark environment, plus each environment's overall goal (\textit{Goal} row).}
\label{tab:s-subtasks}
\small
\renewcommand{\arraystretch}{1.25}
\begin{tabular}{cc p{9.5cm}}
\toprule
\textbf{Env.} & \textbf{\#} & \textbf{Sub-task Definition} \\
\midrule
\multirow{4}{*}{Cabinet} & 1 & Move within 20\,cm of the handle \\
 & 2 & Move within 10\,cm of the handle \\
 & 3 & Open the drawer by 20\,cm \\
 & 4 & Open the drawer by 38\,cm (full task success at 39\,cm) \\
 & \textit{Goal} & \textit{Discrete: open the drawer 39\,cm (one step beyond sub-task 4)} \\
\midrule
\multirow{3}{*}{Factory} & 1 & Align the nut with the bolt \\
 & 2 & Engage the nut with the screw threads (gap $<0.5\times$ thread pitch) \\
 & 3 & Rotate the nut down 0.75\,mm from the engaged position (gap $<0.375\times$ thread pitch; same threshold as full task success) \\
 & \textit{Goal} & \textit{Discrete: same threshold as sub-task 3 (gap $<0.375\times$ thread pitch)} \\
\midrule
\multirow{4}{*}{Lift} & 1 & Reach within 10\,cm of the cube \\
 & 2 & Grasp the cube (within 2\,cm, gripper clamped) \\
 & 3 & Lift the cube 10\,cm into the air \\
 & 4 & Hold the cube within 5\,cm of the elevated target pose \\
 & \textit{Goal} & \textit{Open-ended: no discrete threshold, keep the cube as close to the target pose as possible until truncation} \\
\bottomrule
\end{tabular}
\renewcommand{\arraystretch}{1.0}
\end{table}

\begin{table}[H]
\centering
\caption{Training hyperparameters for each benchmark environment. All values are taken from the default NVIDIA IsaacLab example configurations unless otherwise noted.}
\label{tab:s-training-hyperparameters}
\small
\renewcommand{\arraystretch}{1.15}
\begin{tabular}{lccc}
\toprule
\textbf{Hyperparameter} & \textbf{Cabinet (RSL-RL)} & \textbf{Lift (RSL-RL)} & \textbf{Factory (RL-Games)} \\
\midrule
Algorithm & PPO & PPO & PPO (A2C Continuous)\\
Training iterations & 2500 & 3000 & 200 epochs\\
Environments & 16384 & 4096 & 512\\
Steps / environment & 16 & 24 & 128 (horizon)\\
Total batch size & 65,536 & 98,304 & 16,384\\
Mini-batch size & 8,192 & 24,576 & 512\\
PPO epochs & 8 & 5 & 4\\
Learning rate & $5\times10^{-4}$ & $1\times10^{-4}$ & $1\times10^{-4}$\\
Learning-rate schedule & Adaptive & Adaptive & Adaptive\\
Discount factor ($\gamma$) & 0.99 & 0.98 & 0.995\\
GAE parameter ($\lambda$) & 0.95 & 0.95 & 0.95\\
Clip parameter ($\epsilon$) & 0.2 & 0.2 & 0.2\\
Value loss coefficient & 1.0 & 1.0 & 2.0\\
Entropy coefficient & 0.0 & 0.006 & 0.0\\
Desired KL / Threshold & 0.008 & 0.01 & 0.008\\
Maximum gradient norm & 1.0 & 1.0 & 1.0\\
Initial action noise & 1.0 & 1.0 & Learned log standard deviation\\
Observation normalization & Yes & No & Yes\\
Value normalization & Yes & No & Yes\\
Advantage normalization & -- & -- & Yes\\
Value clipping & Yes & Yes & Yes\\
Reward scaling & Default & Default & 1.0\\
Action clipping & Default & Default & 1.0\\
Mixed precision & Default & Default & Yes\\
\midrule
\multicolumn{4}{c}{\textbf{Policy Network}}\\
\midrule
Hidden layers & [256,128,64] & [256,128,64] & [512,128,64]\\
Activation & ELU & ELU & ELU\\
Shared actor-critic & No & No & Yes\\
\midrule
\multicolumn{4}{c}{\textbf{Recurrent Module}}\\
\midrule
Recurrent network & None & None & LSTM\\
Hidden size & -- & -- & 1024\\
Layers & -- & -- & 2\\
Layer normalization & -- & -- & Yes\\
Concatenate input & -- & -- & Yes\\
\bottomrule
\end{tabular}
\renewcommand{\arraystretch}{1.0}
\end{table}

\section{Extended discussion material}
\label{sec:s4-discussion}

\subsection{Algorithm design history: from a fixed scheduler to ACES}
\label{sec:s4-algo-history}
Supports Section~\ref{sec:distribution-algorithms} of the main paper, expanding the compressed design-history sentence in Section~7.1. Three mechanisms for combining confident and exploratory sampling were evaluated, each addressing a limitation in the one before it. A fixed scheduler, exploratory (softmax) sampling for the first one-fourth of training, then a permanent switch to confident (argmax) sampling, is too blunt: the appropriate switching point depends on how clearly a bottleneck stands out, which varies across seeds and over training, not on a fixed iteration count. The Adaptive Confidence Threshold Sampler (ACTS) instead applies a hard threshold on the margin $m$ (Section~\ref{sec:distribution-algorithms}), switching between exploratory and confident sampling once the confidence distribution becomes sufficiently sparse; a hard threshold, however, still discards information at the boundary itself, treating two margins just barely on either side of the threshold as maximally different despite being nearly identical. ACES resolves this with the continuous blend $\beta$ of Section~\ref{sec:distribution-algorithms}: decisive when a bottleneck is clear, neutral when dense, and continuous in between rather than switching abruptly.

\subsection{Why success rate instead of completion time?}
\label{sec:s4-success-rate-history}
Provides the full design history behind this paper's use of success rate as the difficulty signal (Section \ref{sec:distribution-algorithms}). An earlier version of the difficulty signal used completion time (how long a sub-task took) rather than success rate. Isolating curriculum from non-curriculum episodes matters even more under this metric: environments replayed to later progression states register instant (0.0) completion times when curriculum episodes are included in the measurement, and some sub-tasks inherently take longer than others regardless of the agent's success rate.

Under this scheme, the progression stored a timestamp of sub-task completion rather than a binary completed/not-completed flag, checked every timestep and logged the moment it changed; the value fed into the distribution calculation via the gap between consecutive sub-tasks' completion times, with an uncompleted sub-task's value set to the maximum episode length.

Success rate replaced completion time as the difficulty signal for several reasons. Completion time is noisy, especially under domain randomization, making it a complicated and unclear indicator of difficulty. It also measures speed rather than competence: agents that explore before completing a sub-task register as more difficult than agents that complete it immediately, despite both ultimately succeeding; speed still reflects capability to some degree, but since performance is ultimately measured by success rate rather than speed, conditioning the curriculum on a different mechanism is inconsistent with that goal. Completion times also differ naturally between sub-tasks, each with its own theoretical lower bound, creating inconsistent ceilings across sub-tasks, whereas success rate reaches full performance regardless of the sub-task. Completion time also rewards a fast, low-accuracy policy over a slower, more consistent one: a sub-task completed in 150 steps with certain success and one completed in 75 steps with 50\% success represent two different notions of performance, and consistency is prioritized here, since success achieved only by exhausting every option does not indicate a learned task.

Success rate solves this by being bounded to a clear 0.0--1.0 range and far more resilient to outliers, since it is binary. It is a more direct way to capture the agent's learning and struggles, and its natural normalization makes it significantly easier to smooth using an EMA.

\subsection{Stability analysis}
\label{sec:s4-stability-analysis}
Supports Section \ref{sec:why-sharpening} of the main paper. When we explored different sharpening hyperparameter values for the exploratory sampler, we found that a value of two, instead of one, resulted in a smaller standard deviation and better overall performance. The performance gain is unsurprising, as it is simply hyperparameter tuning, but the change in standard deviation reveals something more interesting about how the curriculum moves between sub-tasks. Intuitively, one might expect a less decisive approach, something more stochastic and less confident, to give more consistent results. However, after observing how this hyperparameter sharpens the curriculum toward larger values, we hypothesize instead that a more rigid, confident distribution can actually be more stable than an indecisive one.

Our working hypothesis is that an unsharpened distribution makes it too easy for the curriculum to drift between sub-tasks: we want the bottleneck to shift from sub-task to sub-task clearly and smoothly as the agent actually resolves each one in turn, and a flatter distribution lets the sampled sub-task wander between several near-equally-weighted options well before the agent has resolved the earlier one. The spread of Cabinet training runs is more dispersed when the sharpening hyperparameter is one rather than two: a less decisive distribution produces less consistent outcomes. This comparison is qualitative, observed during development rather than from a formal $p{=}1$ pure-exploratory ablation, and should be read as a design rationale rather than a statistically tested claim.

While sharpening is not a major driver of end performance, since the confident sampler dominates most of training after the initial startup period, the fact that a more confident sampling method can be more consistent than a neutral one illustrates the algorithm's influence on stability, not just performance.
This connects to a broader pattern in the curriculum's stability story: the Delay ablation (Section \ref{sec:delayed-onset}) is, independently, the lowest-variance curriculum-enabled configuration tested, and Section \ref{sec:instability-replay-specialization}'s diagnostic evidence similarly ties well-performing seeds to more decisive, settled behavior (falling entropy, an early-resolving value-loss spike) rather than continued search. Taken together, decisiveness, whether from the sharpening exponent, from withholding the curriculum early, or from an individual seed's own training dynamics, repeatedly co-occurs with stability in this study, though we stop short of claiming a single unifying mechanism across all three.

\subsection{Determinism validation}
\label{sec:s-determinism}
Supports Section \ref{sec:baselines} and the Limitations' Implementation discussion. Every seed in this study is deterministic: disabling the curriculum reproduces an unmodified, curriculum-free build exactly, and setting the sampling ratio $\gamma=0$ reproduces the disabled-curriculum baseline exactly, confirming the curriculum introduces no unintended side effects when inactive.

This validation process is what first surfaced a subtle pitfall for anyone reimplementing this in a manager-based IsaacLab environment: registering the sub-task progress tracker as an interval-mode event term causes IsaacLab's EventManager to draw from the global RNG every step regardless of the term's own logic, silently desynchronizing curriculum-disabled baselines. We resolved this by implementing the tracker as a zero-returning reward term instead, which carries no hidden RNG draw. This is distinct from eval contamination (Section \ref{sec:overview}), which biases the signal itself rather than the RNG stream, but both are correctness pitfalls that only surface under careful baseline validation like the determinism check above.

\subsection{Delayed curriculum onset (full mechanistic reasoning)}
\label{sec:s4-delayed-onset-full}
Supports Section \ref{sec:delayed-onset} of the main paper.

\paragraph{What delaying the curriculum would change.} Mechanically, disabling the curriculum for the first 15\% of training only changes where resets go: every environment resets to the default pose during this period, rather than a fraction of them drawing from the success buffer as usual. It does not pause anything else. The progression and success buffer continue to populate exactly as described in the main paper's Overview section (Section \ref{sec:overview}), since sub-task completion is tracked from the environment's ongoing rollout regardless of how an episode started, not from whether that episode happened to be a curriculum episode. The practical effect is that by the time the curriculum activates, it would not be starting cold: the success buffer should already hold a reasonable population of world states for each sub-task the agent has reached, and the success-rate signal $r$ driving the distribution algorithm would reflect a policy that has already had time to mature, rather than the first noisy handful of episodes.

\paragraph{Proposed mechanism.} This connects to two ideas discussed elsewhere in the main paper. Exposing the policy to later sub-task resets before it has been trained on the preceding stages is hypothesized to contribute to the curriculum's instability, and difficulty estimation is least reliable early in training, when the current bottleneck is still being resolved from noisy signal. Both problems are worst at the very start of training, before the policy has developed baseline competence and while the success-rate signal driving the curriculum remains uninformative, exactly the window Delay removes. By the time replay begins, a reasonably general policy learned from the default pose alone is expected to already be in place, making concentrated practice on any single sub-task less likely to be the policy's first exposure to that part of the task. This is hypothesized to lower the catastrophic-run rate specifically: replay specialization describes convergence to a narrow, non-generalizing solution for a sub-task, which is more likely when replay constitutes the first exposure to that sub-task than when a general strategy is already available as a fallback.

This finding should not be overstated. Delayed curriculum onset was not a primary hypothesis of this work and is not its flagship contribution; ACES and the empirical bottleneck-targeting curriculum remain that. It was also only tested on the Cabinet environment; without evaluation on Factory or Lift, whether the benefit generalizes across tasks or is specific to Cabinet's particular instability profile remains open. Still, a simple, essentially free change (waiting before enabling the curriculum) is a promising direction this work opens, and warrants dedicated follow-up study in its own right: how much of the curriculum's instability in general is attributable to premature intervention, and how far a delayed-onset strategy can be extended before its cost exceeds its benefit.

Delay has a higher raw mean than the main curriculum (0.7198 vs.\ 0.7032) but a \emph{lower} IQM (0.7647 vs.\ 0.7995), and its paired win rate against ACES is only $31.25\%$ (95\% CI $[0.125, 0.563]$, Table \ref{tab:s-ablation-significance}). These three robust statistics do not all agree with each other, which means Delay's aggregate mean advantage is concentrated in a subset of seeds rather than reflecting a consistent per-seed edge over ACES's typical (trimmed) performance. We do not read this as evidence Delay is unconditionally better than ACES, only that it is a promising, incompletely understood direction worth exploring further as we work on resolving the instability we found in Franka Cabinet.

\subsection{Design guidelines: choosing sub-tasks}
\label{sec:s4-design-guidelines-subtasks}
Supports Section \ref{sec:design-guidelines} of the main paper.

The curriculum's core idea is to let an agent's own empirical, per-sub-task performance decide where its experience goes, rather than fixing that allocation in advance; sub-tasks are what make that decision possible, since they are the interface through which the curriculum measures per-sub-task success and locates the current bottleneck. They decompose the task into checkpoints whose individual success rates the curriculum can measure and target, rather than leaving it only a single, uninformative pass/fail signal for the whole task. Because sub-tasks determine both where training failure becomes visible and where the curriculum resets as a checkpoint, sub-task choice substantially affects curriculum performance. Three and four sub-tasks were used across the environments in this study, and more complex tasks likely require more, though this has not been tested beyond the three-and-four-sub-task environments studied here. In fact, the margin calculation may need to consider the entire dataset (not exclusively the top 2 values) as the sub-task count grows larger. Tasks that cannot be broken down into two or more sub-tasks cannot meaningfully use this curriculum; extending it to less clearly defined sub-tasks, such as those in locomotion tasks, is a direction for future work.

Sub-tasks are an empirical interface, not a claim about the task's true skill decomposition; in practice, ones corresponding to a recognizable skill (e.g.\ touching the cabinet handle, lifting the cube into the air) perform best, and should be well spread out and evenly spaced for the curriculum to handle. When sub-tasks are too difficult to complete, or when important skills are missing between them, the curriculum does not necessarily hurt training directly, but it becomes ineffective, and can indirectly hurt training by diverting curriculum-assigned environments away from contributing to overall learning while they fail to progress. Verifying that all sub-tasks are achievable is recommended, since an infeasible threshold can otherwise be set unintentionally. Placing the final sub-task one small step short of full task completion rather than at its exact threshold is recommended, though not required: in Cabinet, the task is considered complete once the drawer is open 0.39 meters, and sub-task four was set to an opening of 0.38 meters, conditioning the policy to learn that final step. The reasoning extends beyond this empirical result: a final sub-task whose success criterion is identical to full-task success cannot provide a distinguishable, earlier bottleneck signal, since its measured success rate would simply track the (typically low) overall success rate rather than something the curriculum can act on before the task is essentially already solved. Factory is a real counter-example in this study, not a hypothetical one: its final sub-task (rotating the nut down to the engaged threshold) uses the same threshold as full task success itself, since the natural physical milestone coincides with completion. Factory's results (Section~\ref{sec:results}) do not show this hurting performance; the curriculum still improves an already-saturated baseline, so this remains a recommendation rather than a confirmed requirement, though Factory's baseline was already near ceiling regardless of sub-task design, which may limit how much a distinguishable bottleneck signal could have helped further.

Exploring the different ways a sub-task can be technically expressed for different robot tasks, evaluating which formulations perform best, and studying the effects of gaps between them are directions for future research.

\subsection{Design guidelines: choosing world representations}
\label{sec:s4-design-guidelines-world-rep}
Supports Section \ref{sec:design-guidelines} of the main paper.

The world representation determines the reset pose the curriculum ultimately produces; it is the mechanism by which the curriculum affects training, converting the difficulty estimate into a concrete environment change. The appropriate world representation generally follows from how the environment itself is structured: moving objects and articulation joints should have their orientations and positions saved. The main design choice is whether to include velocities (joint, angular, or linear). More faithfully replicating the state at which a sub-task was completed is expected to let training continue more naturally into the next sub-task, producing a more consistent curriculum; this holds for positions and joints, since omitting them would leave the reset an incomplete replay. Whether to also include the object's velocities at the moment of completion remains an open question.

This was tested in the Franka Cabinet environment through the Velocity ablation (Appendix~\ref{sec:s1-ablations} above), which evaluated whether including velocities improves training performance; it did not. Why additional data would degrade performance remains speculative, though a plausible explanation follows from causal consistency: the stored velocity for the drawer being opened is not produced by any preceding action of the agent's, and is present immediately upon reset, decoupled from the policy's own action history. Replaying it therefore functions as an uncaused, unpredictable signal rather than evidence of motion the agent is in the process of executing, which may account for the resulting drop in performance. This explanation is not confirmed and warrants further investigation; it questions whether more data is always beneficial.

\section{Full keywords glossary}
\label{sec:s5-keywords}

This is the paper's complete glossary: it reproduces the main paper's Keywords section (Section~\ref{sec:keywords}) in full below, then covers additional terminology, including the curriculum-specific diagnostics referenced in Appendix~\ref{sec:s2-diagnostics}, the statistical tests used throughout the Results and Appendix~\ref{sec:s1-ablations}, and general keywords for the reinforcement learning field.

\textbf{Sub-task} Intermediate milestone toward the overarching task. \newline
\textbf{Bottleneck} Earliest sub-task where progress is currently most stalled; the curriculum's replay target. \newline
\textbf{World State} An environment's configuration (positions, velocities, orientations, and joint state) captured at sub-task completion and sufficient to reset the simulator back into it. \newline
\textbf{Progression} Live, per-environment tensor of sub-task completion and world state. \newline
\textbf{Success Buffer} FIFO memory of world states saved at each sub-task's completion; sampled from. \newline
\textbf{Sampling Ratio} Fraction of resets that become \emph{curriculum episodes} rather than the environment's default (\emph{non-curriculum}) pose. \newline
\textbf{Difficulty} Complement of (EMA-smoothed) success rate. \newline
\textbf{Confident Sampling} Selecting the highest-difficulty sub-task via $\operatorname{argmax}$. \newline
\textbf{Exploratory Sampling} Sampling a sharpened difficulty distribution via $\operatorname{softmax}$. \newline
\textbf{ACES} Adaptive Confidence-Exploration Sampler, continuously blending confident and exploratory sampling by margin. \newline
\textbf{Replay Specialization} A narrow, non-generalizing solution learned from replayed resets that fails to transfer to the full task. \newline

\textbf{Reset Distance and Variance} An averaged measure of how far reset poses drift from the default position, and how much that distance varies across environments (Appendix~\ref{sec:s2-diagnostics}). \newline
\textbf{Selected} The sub-task index most recently targeted by curriculum resets (Appendix~\ref{sec:s2-diagnostics}). \newline
\textbf{Difficulty ($d$)} The failure rate, or complement of the EMA-smoothed success rate, representing how hard a sub-task currently is for the agent (Appendix~\ref{sec:s2-diagnostics}). \newline
\textbf{Gap ($g$)} The difficulty differential between a sub-task and the one before it; the raw signal ACES samples from to find the earliest bottleneck (Appendix~\ref{sec:s2-diagnostics}). \newline
\textbf{Sampling Distribution} ACES's resulting per-sub-task selection probability (Appendix~\ref{sec:s2-diagnostics}). \newline
\textbf{Realized Sampling Rate} The actual fraction of resetting environments that become curriculum episodes, which should track the configured target ratio (0.3 for the main curriculum run; Appendix~\ref{sec:s2-diagnostics}). \newline

\textbf{Interquartile Mean (IQM)}~\cite{agarwal2021deep} The mean of the middle 50\% of a sorted set of values (i.e.\ the mean after discarding the bottom and top quartile); this paper's primary summary statistic, chosen for its robustness to outliers relative to the raw mean. \newline
\textbf{Wilcoxon (Signed-Rank Test)} A paired, non-parametric significance test used throughout this paper's tables (e.g.\ Table~\ref{tab:math-analysis}) to assess whether an experimental condition's per-seed values differ systematically from baseline's. \newline
\textbf{$P(X{>}Y)$} Probability of superiority: the probability that a randomly drawn value from the experimental condition ($X$) exceeds a randomly drawn value from the baseline condition ($Y$), estimated empirically as the fraction of same-seed pairs where the experimental value wins. Reported in this paper's tables via its 95\% bootstrap CI (below) rather than as a bare point estimate. \newline
\textbf{95\% Bootstrap CI}~\cite{agarwal2021deep} This paper's fourth paired-comparison statistic (alongside IQM, Cliff's $\delta$, and the Wilcoxon test): a 95\% confidence interval on $P(X{>}Y)$, computed by resampling the $n$ same-seed pairs with replacement $B=10{,}000$ times, recomputing $P(X{>}Y)$ on each resample, and taking the 2.5th/97.5th percentile of the resulting distribution. Reported in place of a bare $P(X{>}Y)$ point estimate throughout Tables~\ref{tab:math-analysis} and \ref{tab:s-ablation-significance}, following Agarwal et al.'s recommendation to report an interval rather than a single number for exactly this statistic. \newline
\textbf{Cliff's $\delta$}~\cite{agarwal2021deep} A non-parametric effect size derived from $P(X{>}Y)$ as $\delta = 2P(X{>}Y) - 1$, ranging from $-1$ (every baseline value exceeds every experimental value) to $+1$ (the reverse). \newline

\textbf{ACTS} Adaptive Confidence Threshold Sampler: switches between confident and exploratory sampling using a hard threshold on the margin, superseded by ACES's continuous blend (Section~\ref{sec:distribution-algorithms}). \newline
\textbf{Curriculum Episode} An episode whose reset pose is selected by the curriculum rather than the environment's default pose. \newline
\textbf{Non-curriculum Episode} An episode reset to the environment's default pose, unaffected by the curriculum; used to measure ground-truth success rate. \newline
\textbf{Eval Contamination} The failure mode where curriculum episodes are inadvertently included when measuring the success-rate signal, biasing the difficulty estimate the curriculum itself depends on. Avoided by computing success rate only from non-curriculum episodes. \newline
\textbf{Success Rate} The EMA-smoothed fraction of non-curriculum episodes that have completed a given sub-task; used to estimate difficulty. \newline
\textbf{Margin ($m$)} The difference between the two largest values of the confidence tensor (Section~\ref{sec:distribution-algorithms}); a large margin indicates one clear bottleneck, while a small margin indicates several similarly difficult sub-tasks. Tracked as a diagnostic over training in Appendix~\ref{sec:s2-diagnostics}. \newline
\textbf{Blend ($\beta$)} The weight ACES assigns to its confident component relative to its exploratory component, determined by the margin (Section~\ref{sec:distribution-algorithms}). Tracked as a diagnostic over training in Appendix~\ref{sec:s2-diagnostics}. \newline
\textbf{Catastrophic Run} A training run whose success rate remains near zero, far below the typical range of otherwise-identical runs. \newline

\textbf{IsaacSim/Lab} A robotics simulation platform developed by NVIDIA that provides realistic physics for GPU-parallelized training of robot agents. \newline
\textbf{Sim-to-Real} The process of deploying a policy trained in a simulation tool (such as IsaacLab) onto a physical robot that must operate in the real world. \newline
\textbf{Terminations} Ending the episode early, generally because of succeeding the task or breaching a safety precaution. \newline
\textbf{Truncations} Ending the episode because of a timeout. \newline
\textbf{Observations} The information given to the agent to determine what action to take. \newline
\textbf{Domain Randomization} Noise added to different aspects of the environment to assist in generalization. \newline
\textbf{Generalized Advantage Estimation (GAE)} The technique PPO uses internally to estimate how much better or worse an action was than expected, trading off bias and variance via a parameter $\lambda$. \newline
\textbf{KL Divergence} A measure of how much a policy has changed between updates; PPO's adaptive learning-rate schedule targets a desired KL divergence to keep updates neither too large nor too small. \newline
\textbf{Entropy (Coefficient)} A measure of how stochastic, or spread out, a policy's action distribution is; higher entropy means more exploration. The entropy coefficient is a PPO hyperparameter that rewards maintaining higher entropy. \newline
\textbf{Manager-Based / Direct} Two ways of structuring an IsaacLab environment: Direct environments implement rewards, resets, and observations in a single class, while Manager-Based environments compose these from modular, reusable manager components. \newline
\textbf{Reward Curriculum} A curriculum that modifies the reward system throughout the training to encourage new behavior. \newline
\textbf{Domain Randomization Curriculum} A curriculum that begins with a narrow, predictable range of randomization and gradually widens it to benefit generalization. \newline
\textbf{Environment Curriculum} A curriculum that changes the environment the agent must interact with to build up to a more difficult interaction. \newline
\textbf{Task} The goal the robot is intended to complete. \newline
\textbf{World Representation} The design choice of which physical quantities (positions, velocities, joint values, etc.) are included in a world state. \newline
\textbf{Distribution} The probability distribution over sub-tasks from which a curriculum episode's target is drawn. \newline
\textbf{Exponential Moving Average (EMA) Distribution} A momentum-based smoothing method that gradually updates the distribution over a set time constant rather than replacing it outright. \newline

\newpage
\section*{NeurIPS Paper Checklist}

\begin{enumerate}

\item {\bf Claims}
    \item[] Question: Do the main claims made in the abstract and introduction accurately reflect the paper's contributions and scope?
    \item[] Answer: \answerYes{}
    \item[] Justification: The abstract and Section~2 (Introduction) state the contribution (a curriculum that adapts reset pose to the agent's own per-sub-task success), the evaluated scope (three Franka manipulation environments, up to 16 seeds each), and report all three environments (including Factory, evaluated at $n=8$ given its longer training time) as completed results rather than overclaiming beyond what was tested.

\item {\bf Limitations}
    \item[] Question: Does the paper discuss the limitations of the work performed by the authors?
    \item[] Answer: \answerYes{}
    \item[] Justification: Section~8 (Limitations) has dedicated Scope and Replay-specialization/implementation discussions covering seed counts, the narrow (three-environment, single-arm) evaluation, no sim-to-real evaluation, the manual sub-task-decomposition requirement, the lack of alternative curriculum benchmarking, and expected training-time overhead.

\item {\bf Theory assumptions and proofs}
    \item[] Question: For each theoretical result, does the paper provide the full set of assumptions and a complete (and correct) proof?
    \item[] Answer: \answerNA{}
    \item[] Justification: The paper is purely empirical; Section~4 gives closed-form equations defining the curriculum algorithm (ACES), but these are algorithmic definitions, not theorems requiring proof.

    \item {\bf Experimental result reproducibility}
    \item[] Question: Does the paper fully disclose all the information needed to reproduce the main experimental results of the paper to the extent that it affects the main claims and/or conclusions of the paper (regardless of whether the code and data are provided or not)?
    \item[] Answer: \answerYes{}
    \item[] Justification: Section~4 fully specifies the curriculum algorithm (equations for difficulty estimation, gap computation, and the ACES blend), Section~5 and Table~1 give the curriculum's own hyperparameters, and Appendix~D gives per-environment sub-task thresholds and full PPO/network training hyperparameters, all sufficient to reproduce the method on the cited IsaacLab example environments. Code is additionally provided for reference in the supplementals.

\item {\bf Open access to data and code}
    \item[] Question: Does the paper provide open access to the data and code, with sufficient instructions to faithfully reproduce the main experimental results, as described in supplemental material?
    \item[] Answer: \answerYes{}
    \item[] Justification: Anonymized source code for the curriculum implementation and the environment-specific sub-task/reset-pose modifications is included as supplementary material (a fully anonymized zip file, per the workshop's supplementary-material policy); the base simulation environments themselves are IsaacLab's public example configurations (cited, Section~5.1).

\item {\bf Experimental setting/details}
    \item[] Question: Does the paper specify all the training and test details (e.g., data splits, hyperparameters, how they were chosen, type of optimizer) necessary to understand the results?
    \item[] Answer: \answerYes{}
    \item[] Justification: Section~5 gives per-environment sub-task definitions, iteration/environment counts, and seed ranges; Table~1 gives the curriculum's own hyperparameters; Appendix~D (Table~7) gives the full PPO/network training configuration (optimizer settings, learning rate, batch sizes, architecture) for each environment.

\item {\bf Experiment statistical significance}
    \item[] Question: Does the paper report error bars suitably and correctly defined or other appropriate information about the statistical significance of the experiments?
    \item[] Answer: \answerYes{}
    \item[] Justification: Table~2 (main text) and Appendix~A report IQM, mean/median/standard deviation/min/max, paired two-sided Wilcoxon signed-rank $p$-values, Cliff's $\delta$ effect sizes, and a 95\% bootstrap confidence interval on $P(X{>}Y)$ ($B=10{,}000$ paired resamples), all computed by seed ($n=16$ for Cabinet/Lift, $n=8$ for Factory) across identical seed pairs between baseline and curriculum; Figure~2's shaded bands are $\pm1$ sample standard deviation across seeds, stated in each caption. Additionally our Appendix details our ablation tests clearly with both visual figures and metrics.

\item {\bf Experiments compute resources}
    \item[] Question: For each experiment, does the paper provide sufficient information on the computer resources (type of compute workers, memory, time of execution) needed to reproduce the experiments?
    \item[] Answer: \answerYes{}
    \item[] Justification: Appendix~D states the hardware used for all experiments (a single workstation: an RTX~5090 GPU with 32\,GB VRAM, a 9950X3D CPU, and 64\,GB RAM); exact per-run wall-clock training time is not reported, but environment/iteration counts (Table~7) let readers estimate relative training cost across environments.
    
\item {\bf Code of ethics}
    \item[] Question: Does the research conducted in the paper conform, in every respect, with the NeurIPS Code of Ethics \url{https://neurips.cc/public/EthicsGuidelines}?
    \item[] Answer: \answerYes{}
    \item[] Justification: The research involves only simulated robot manipulation training (IsaacLab); it uses no human subjects, no personal or scraped data, and no dual-use or high-risk assets, and the author(s) have reviewed the NeurIPS Code of Ethics and identify no deviation.

\item {\bf Broader impacts}
    \item[] Question: Does the paper discuss both potential positive societal impacts and negative societal impacts of the work performed?
    \item[] Answer: \answerNo{}
    \item[] Justification: The paper does not include a dedicated societal-impact discussion. The contribution is a training-time curriculum for simulated manipulation policies, evaluated only in simulation with no deployment described; any downstream societal impact would be inherited from the specific robot/policy it is later applied to, not from the curriculum itself, consistent with this being foundational research not tied to a particular application.
    
\item {\bf Safeguards}
    \item[] Question: Does the paper describe safeguards that have been put in place for responsible release of data or models that have a high risk for misuse (e.g., pre-trained language models, image generators, or scraped datasets)?
    \item[] Answer: \answerNA{}
    \item[] Justification: The paper releases no data, pretrained models, or other assets with a high risk for misuse; it describes a training-time curriculum algorithm evaluated in simulation only.

\item {\bf Licenses for existing assets}
    \item[] Question: Are the creators or original owners of assets (e.g., code, data, models), used in the paper, properly credited and are the license and terms of use explicitly mentioned and properly respected?
    \item[] Answer: \answerYes{}
    \item[] Justification: The paper cites the original sources of all software assets used (IsaacLab~\cite{mittal2025isaaclab}, RSL-RL~\cite{schwarke2025rslrl}, PPO~\cite{schulman2017proximal}, RL-Games) in Section~5.1, and Appendix~D states each asset's license (IsaacLab and RSL-RL: BSD-3-Clause; RL-Games: MIT License).

\item {\bf New assets}
    \item[] Question: Are new assets introduced in the paper well documented and is the documentation provided alongside the assets?
    \item[] Answer: \answerYes{}
    \item[] Justification: The paper introduces no new datasets or models, but does release new code: the curriculum implementation itself (an algorithmic modification layered onto existing IsaacLab benchmark environments) is included as an anonymized zip in the supplementary material, documented with instructions covering installation, exact training commands per environment, and the curriculum hyperparameters.

\item {\bf Crowdsourcing and research with human subjects}
    \item[] Question: For crowdsourcing experiments and research with human subjects, does the paper include the full text of instructions given to participants and screenshots, if applicable, as well as details about compensation (if any)?
    \item[] Answer: \answerNA{}
    \item[] Justification: The paper involves no crowdsourcing or human subjects; all experiments are simulated robot reinforcement learning.

\item {\bf Institutional review board (IRB) approvals or equivalent for research with human subjects}
    \item[] Question: Does the paper describe potential risks incurred by study participants, whether such risks were disclosed to the subjects, and whether Institutional Review Board (IRB) approvals (or an equivalent approval/review based on the requirements of your country or institution) were obtained?
    \item[] Answer: \answerNA{}
    \item[] Justification: The paper involves no human subjects or crowdsourcing; no IRB review applies.

\item {\bf Declaration of LLM usage}
    \item[] Question: Does the paper describe the usage of LLMs if it is an important, original, or non-standard component of the core methods in this research? Note that if the LLM is used only for writing, editing, or formatting purposes and does \emph{not} impact the core methodology, scientific rigor, or originality of the research, declaration is not required.
    \item[] Answer: \answerNA{}
    \item[] Justification: LLM assistance was used only to edit, revise, and format the paper (including generating figures/analysis scripts from the authors' own recorded experimental logs); the curriculum algorithm (Section~4) and experimental design are the authors' own original work, and LLM usage did not impact the core methodology, scientific rigor, or originality of the research.

\end{enumerate}

\end{document}